\documentclass[preprint,12pt,authoryear]{elsarticle}

\usepackage{amssymb}
\usepackage{amsmath}
\usepackage{amsmath}
\usepackage{graphicx}
\usepackage[colorlinks=true, allcolors=blue]{hyperref}
\usepackage{listings}
\usepackage{algorithm}
\usepackage{algpseudocode}
\usepackage{enumitem}
\usepackage{booktabs}
\usepackage{multirow}
\usepackage{multicol}
\usepackage{soul}
\usepackage{float}
\usepackage{doi}

\newcounter{appendixaalgorithm}

\newcounter{appendixbtable}

\newcounter{appendixdlisting}

\newcounter{appendixetable}

\newcounter{appendixflisting}

\newcounter{appendixglisting}

\journal{Transportation Research Part C: Emerging Technologies}

\begin{document}

\begin{frontmatter}



\title{GATSim: Urban Mobility Simulation with Generative Agents}


\author[label1]{Qi Liu}
\author[label1]{Can Li}
\author[label1]{Wanjing Ma}

\affiliation[label1]{organization={Key Laboratory of Road and Traffic Engineering of the Ministry of Education, College of Transportation, Tongji University},
            addressline={4800 Cao'an Rd.}, 
            city={Shanghai},
            postcode={201804}, 
            state={Shanghai},
            country={P.R. China}}

\begin{abstract}
Traditional agent-based urban mobility simulations often rely on rigid rule-based systems that struggle to capture the complexity, adaptability, and behavioral diversity inherent in human travel decision making. Inspired by recent advancements in large language models and AI agent technologies, we introduce GATSim, a novel framework that leverages these advancements to simulate urban mobility using generative agents with dedicated cognitive structures. GATSim agents are characterized by diverse socioeconomic profiles, individual lifestyles, and evolving preferences shaped through psychologically informed memory systems and lifelong learning. The main contributions of this work are: 1) a comprehensive architecture that integrates urban mobility foundation model with agent cognitive systems and transport simulation environment; 2) a hierarchical memory designed for efficient retrieval of contextually relevant information, incorporating spatial and temporal associations; 3) planning and reactive mechanisms for modeling adaptive mobility behaviors which integrate a multi-scale reflection process to transform specific travel experiences into generalized behavioral insights. Experiments indicate that generative agents perform competitively with human annotators in role-playing scenarios, while naturally producing realistic macroscopic traffic patterns. The code for the prototype implementation is publicly available at \url{https://github.com/qiliuchn/gatsim}.
\end{abstract}


\end{frontmatter}


\section{Introduction}
Agent-based modeling (ABM) is a powerful tool for simulating complex systems, including urban traffic, epidemic spread, and market dynamics \citep{macal2016everything}. In the domain of urban planning and traffic management, urban mobility simulation systems play a crucial role by enabling researchers and policymakers to model complex travel patterns and evaluate infrastructure changes \citep{bastarianto2023agent}. While existing systems like \texttt{MATSim} \citep{matsim_examples} have demonstrated significant capabilities in modeling large-scale urban mobility flows, they face fundamental limitations in capturing the heterogeneity, complexity and adaptability of human travel behavior \citep{huang2022overview}. 

Recent advances in large language models (LLMs) offer a transformative opportunity to address these limitations. The pioneering work \citet{park2023generative} introduced ``generative agents'' to demonstrate the feasibility of using LLMs to create more realistic simulation agents. However, existing generative agent studies fall short in addressing the complexities of urban mobility decision-making. Their general-purpose cognitive architectures are not well-suited to producing realistic travel behaviors, as they typically lack adaptive learning mechanisms essential for simulating human-like adaptability in urban mobility scenarios.

Given these limitations, there is a clear need for a more specialized approach to model the dynamics of urban mobility. To this end, we introduce \textit{GATSim (Generative-Agent Transport Simulation)}, a novel framework designed to leverage the strengths of LLMs while addressing the challenges of urban mobility modeling. In contrast to previous studies on generative agents, we customize the memory, planning, reaction and reflection mechanisms specifically for better modeling urban activities. GATSim agents possess diverse socioeconomic attributes, individual lifestyles, and evolving preferences that shape their mobility decisions through psychologically-informed memory systems and lifelong learning mechanisms. This paper makes the following contributions to the field of urban mobility simulation:

\begin{enumerate}
\item \textbf{A comprehensive framework for generative-agent-based urban mobility simulation}: We propose GATSim, a novel framework that integrates urban mobility foundation model with agent cognitive structures and transport simulation environment.

\item \textbf{A memory system designed for efficient retrieval of contextually relevant information}: We develop a memory architecture that enables efficient retrieval of spatially and temporally associated information. The system incorporates multi-modal retrieval mechanisms combining keyword matching, semantic similarity, and spatial-temporal relevance to support complex agent reasoning in urban mobility scenarios.

\item \textbf{Innovative planning and reactive mechanisms for modeling adaptive mobility behaviors}: We introduce novel planning and reactive mechanisms that enable agents to dynamically adapt to evolving urban environments while ensuring lifelong learning capabilities. These mechanisms leverage multi-scale reflection processes to systematically transform individual travel experiences into generalized behavioral patterns.
\end{enumerate}

\begin{figure}
    \centering
    \includegraphics[width=0.95\linewidth]{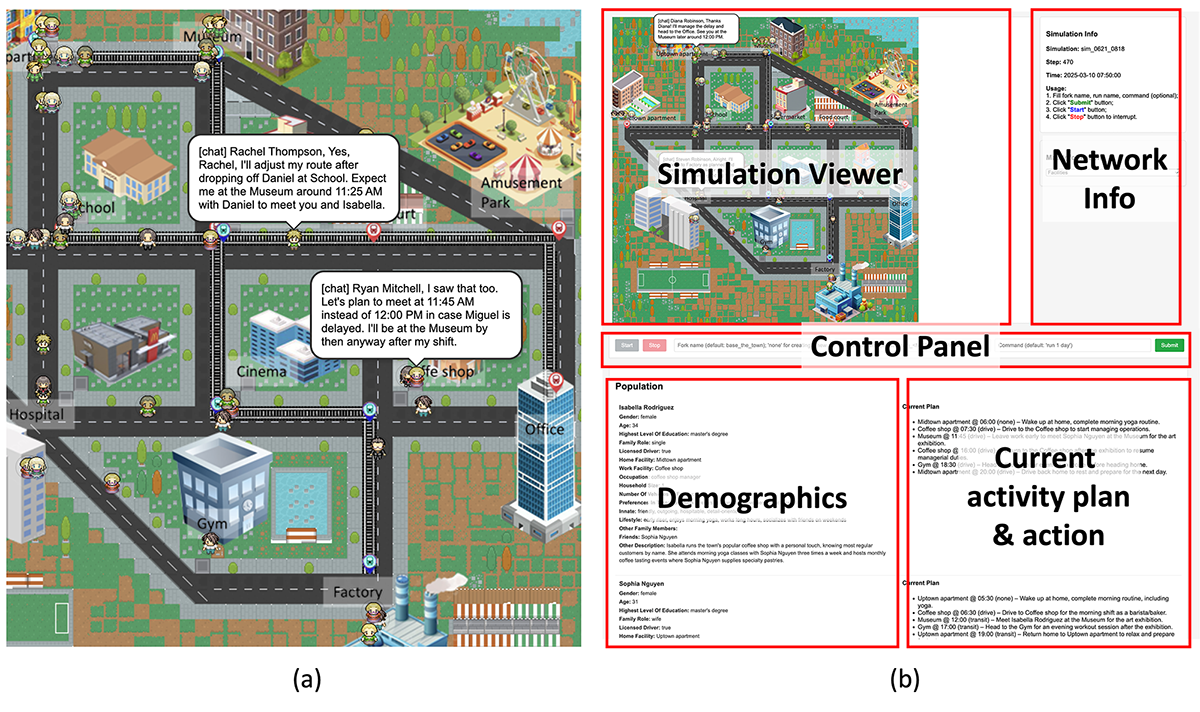}
    \caption{GATSim prototype implementation. (a) Generative agent activity revision and en-route decision making. It demonstrates how agents dynamically adapt their activity plans and make real-time travel decisions based on current conditions, retrieved memories, and contextual information. (b) GATSim Prototype GUI. Users input the \texttt{fork name}, \texttt{run name}, and \texttt{command}, then click ``submit'' and ``start'' to initiate simulation. The interface displays network information, demographic data, and real-time agent plans and actions.}
    \label{fig:intro}
\end{figure}

A prototype of GATSim framework is implemented (See Figure~\ref{fig:intro}). To validate GATSim at both microscopic (i.e., individual agent behaviors) and macroscopic (i.e., overall traffic patterns) levels, we conduct a series of experiments. Human-agent behavioral comparison studies is conducted using LLM-as-judge to assess the realism and adaptability of GATSim agents. Results show that GATSim agents perform competitively with human annotators and they naturally generate realistic macroscopic traffic patterns without explicit behavior coding. Through this comprehensive methodology, GATSim contributes to the paradigm shift from rule-based to intelligence-based urban mobility simulation, providing a more realistic, flexible, and user-friendly framework for urban mobility modeling and analysis. Table~\ref{tab:framework_comparison} highlights key differences between GATSim and traditional agent-based mobility simulation frameworks (e.g., MATSim).

\begin{figure}
    \centering
    \includegraphics[width=0.85\linewidth]{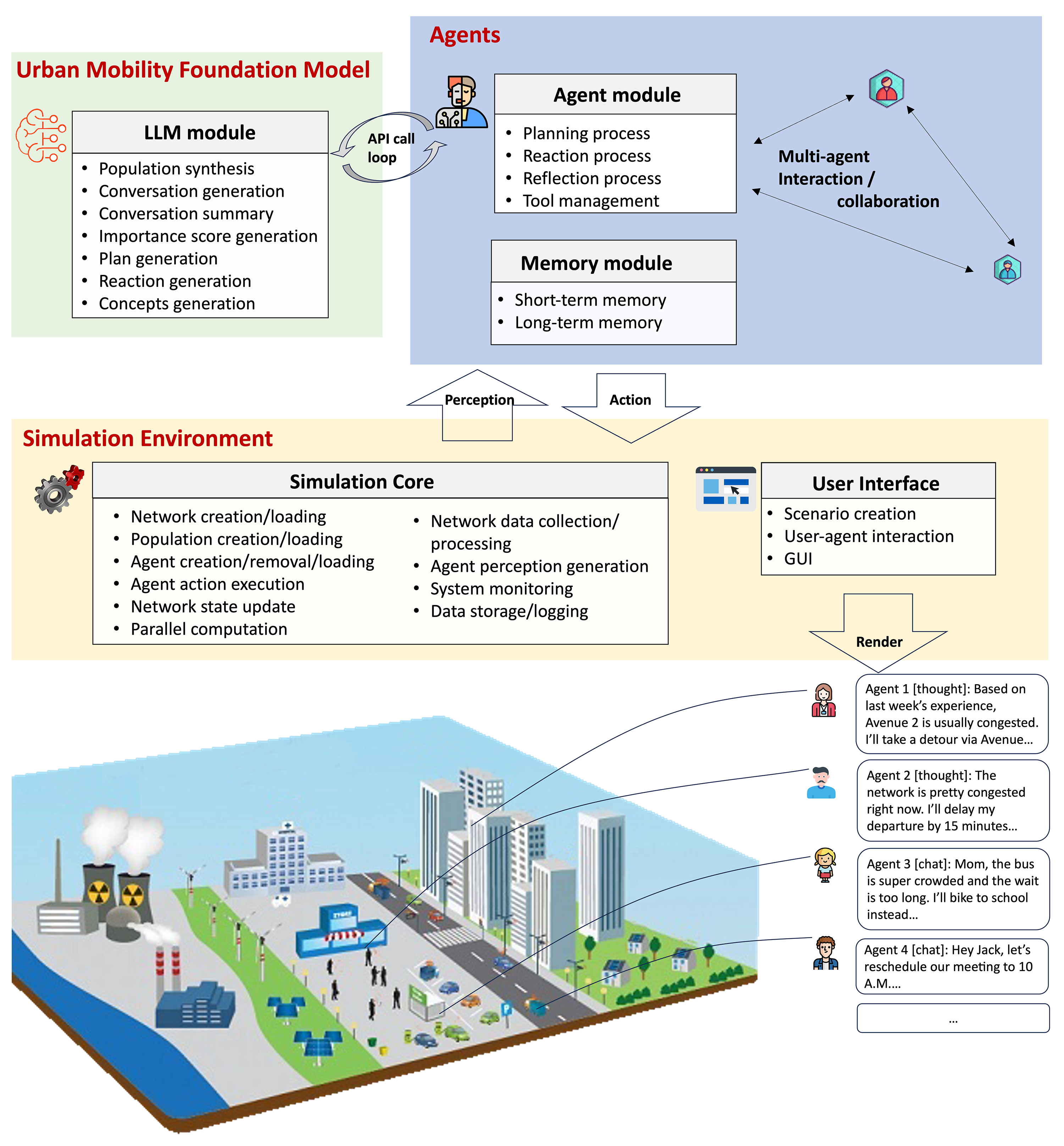}
    \caption{Overall architecture of the GATSim framework, showing the interaction between the Urban Mobility Simulation Environment and the Generative Agents Module. The framework integrates three core components: the Urban Mobility Foundation Model for population synthesis and behavior generation, generative agents with cognitive architectures designed for urban mobility, and the simulation environment for system dynamics and user interaction management.}
    \label{fig:framework}
\end{figure}

GATSim integrates three core components: a simulation environment that manages system dynamics (Sec~\ref{sec:environment}), an urban mobility foundation model (Sec~\ref{sec:llm}), and generative agents with cognitive architectures (Sec~\ref{sec:agents}), as illustrated in Figure~\ref{fig:framework}. This modular design enables flexible integration of different LLM and simulation environments. Framework components will be introduced sequentially. Then Sec~\ref{sec:experiments} presents the experiments. Sec~\ref{sec:conclusion} concludes.

\begin{table}[h]
\centering
\caption{Comparison of GATSim with traditional simulation frameworks}
\label{tab:framework_comparison}
\begin{tabular}{p{4.2cm}p{4cm}p{4cm}}
\toprule
\textbf{Capability} & \textbf{Traditional} & \textbf{GATSim} \\
\midrule
Behavioral Representation & Rule-based & LLM-powered \\
Adaptation Mechanism & Fixed or optimization & Reflective learning \\
Social Interaction & Scripted or absent & Natural language \\
Scenario Customization & Manual coding required & Natural language input \\
Memory System & Limited state variables (eg. set of activity chains) & Hierarchical memory system storing experiences, plans, and reflections\\
Traveler Reasoning Process & Opaque or absent & Transparent and interpretable \\
\bottomrule
\end{tabular}
\end{table}

\section{Related Work} 
\label{sec:related work}

Transportation simulation methodologies span a spectrum from macroscopic approaches that model aggregate traffic flows to microscopic frameworks that capture individual movements and behaviors. Agent-based modeling represents a particularly powerful microscopic approach where autonomous entities make independent decisions based on local information and behavioral rules. This section first reviews recent applications of LLMs in transportation and urban systems; then reviews the evolution of transportation simulation studies, beginning with traditional macroscopic systems, then examining recent advances in LLM-based agent simulation across various domains, and concluding with an analysis of current research gaps that motivate our work.

\subsection{LLMs in Transportation and Urban Systems}

The transportation domain has seen growing interest in LLM applications, spanning prediction, simulation, and control tasks. Traffic prediction systems like UrbanGPT \citep{li2024urbangpt} and OpenCity \citep{li2024opencity} employ spatiotemporal attention mechanisms for city-scale forecasting, while specialized models like TransGPT \citep{wang2024transgpt} and TrafficGPT \citep{zhang2024trafficgpt} focus on transportation control tasks. For urban simulation and behavioral modeling, several approaches have emerged. \citet{xu2023urban} proposed Urban Generative Intelligence (UGI), incorporating city-specific foundation models for various urban tasks. \citet{jin2023surrealdriver} combined LLM agents with the CARLA simulator for microscopic driving behavior, while \citet{yuan2023learning} developed frameworks based on psychological theories like Maslow's hierarchy of needs.

Recent work has specifically targeted mobility behavior generation. \citet{shao2024chain} introduced Chain-of-Planned-Behaviour (CoPB) for structured mobility generation, while \citet{wang2024simulating} developed desire-driven frameworks for activity trajectory synthesis. \citet{li2024more} proposed Desire-driven Autonomous Agents (D2A) for travel diary generation, and \citet{wang2024large} introduced self-consistency approaches to align LLMs with real-world activity data. \citet{bhandari2024urban} focused on synthetic travel survey generation using LLM-based systems. Several studies have explored mobility prediction using historical data. Research by \citet{wang2023would, gong2024mobility, chen2025toward} focuses on leveraging location-based service (LBS) data and mobility histories to predict future travel patterns, demonstrating the potential of LLMs for understanding spatial-temporal mobility dynamics. These diverse applications highlight the growing recognition of LLMs as powerful tools for modeling complex urban and transportation systems.

\subsection{Traditional Urban Mobility Simulation Methods}

Urban mobility simulation methodologies have evolved from macroscopic to increasingly microscopic approaches over several decades. The Four-Step Model \citep{ortuzar2002modelling} represents one of the earliest macroscopic approaches. Travel-chain-based modeling emerged in the 1980s and 1990s \citep{kitamura1996applications}, explicitly modeling sequences of trips individuals make throughout a day rather than treating trips in isolation. Activity-based modeling became particularly influential starting in the mid-1990s \citep{bowman1995activity, axhausen1992activity}, expanding the travel-chain concept by focusing on individual activities, their scheduling, duration, location, and sequence. Landmark implementations including the Day Activity Schedule model \citep{bowman2001activity} and \citet{arentze2004learning}. More recent studies such as \citet{galland2014multi}, \citet{marilleau2006agent}, \citet{bucchiarone2019collective}, and \citet{li2021systematic} adopt more advanced expert models and explore a wider range of scenarios, including emergency excavation, carpooling, and autonomous vehicle simulation. Agent-based simulation gained widespread adoption in urban traffic modeling from the late 1990s onwards, driven by advances in computational power and influential frameworks such as TRANSIMS \citep{smith1995transims} and MATSim \citep{horni2016matsim}. Contemporary platforms including AIMSUN \citep{casas2010traffic} and CUBE \citep{wong2008transportation} represent mature implementations of these methodologies. 

Traditional agent-based modeling approaches primarily rely on rule-based systems, optimization methods, or reinforcement learning. Rule-based approaches remain prevalent due to their interpretability, while optimization-based methods enable agents to maximize utility or reward functions \citep{charypar2005generating, horni2016matsim}. Recent innovations have extended these approaches through deep learning integration \citep{wong2018discriminative} and activity pattern recognition from trajectory data \citep{adetiloye2019multimodal, lou2020detecting}. However, traditional methods fundamentally limit the realism of human behavior modeling due to their inability to capture behavioral diversity, their reliance on rigid and non-adaptive decision-making, poor generalization across contexts, and the absence of mechanisms for social interaction and learning.

\subsection{LLM-Based Agent Simulation and Platforms}

Recent advances in LLMs have enabled the development of generative agents—autonomous entities that produce human-like behaviors through natural language reasoning and memory-based learning. The seminal work by \citet{park2023generative} established the foundational architecture. In economics and finance, agents support macroeconomic simulation \citep{li2023econagent, horton2023large} and financial market modeling \citep{gao2024simulating}. Social science applications include opinion polling \citep{park2024generative}, crisis management \citep{xiao2023simulating}, and information diffusion analysis \citep{zhang2024large, rossetti2024social}. Game theory research explores competitive and cooperative agent behaviors \citep{xie2024can, guo2024economics, fontana2024nicer}. These successes across varied contexts demonstrate the broad applicability of LLM-based agent simulation.

Large-scale multi-agent platforms have emerged to support complex simulations at scale. AgentScope \citep{gao2024agentscope} and OASIS \citep{yang2024oasis} focus simulations with thousands to millions of interacting agents, demonstrating technical feasibility of large-scale deployment. These platforms potentially can provide robust computational infrastructure for agent coordination and parallel execution.

\subsection{Research Gaps and Opportunities}

Despite significant progress, existing approaches face critical limitations that prevent comprehensive urban mobility simulation:

\textbf{Fragmented Focus}: Current studies predominantly address isolated aspects such as next location prediction, lacking integrated frameworks that combine realistic population synthesis, behavioral modeling, and system dynamics.

\textbf{Limited Behavioral Complexity}: Existing systems fail to capture the full spectrum of human decision-making processes, including spatial-temporal memory utilization, adaptive learning from experience.

\textbf{Inadequate Social Dynamics}: Most approaches treat agents as independent entities, missing the collaborative coordination, information sharing, and social influence that characterize real urban mobility systems.

Recent advances in generative agents \citep{park2023generative} and large-scale multi-agent systems \citep{gao2024agentscope, yang2024oasis} offer new opportunities to bridge these gaps. In this work, we present a comprehensive framework that integrates advanced agent cognition with a scalable simulation architecture, enabling realistic behavioral diversity while ensuring computational efficiency for practical applications.

\section{Simulation Environment} 
\label{sec:environment}

The simulation core manages the fundamental computational processes including transportation network management, population management and simulation loop execution. The key distinguishing feature of the proposed simulation core is its capability to generate individualized perceptions for agents and execute the plan-action-reflection cognitive loop that enables behavioral learning and adaptation.

In the prototype, the simulation core is designated as \texttt{backend} module. The \texttt{backend} serves as the simulation entry point, loading environment configuration from \texttt{gatsim/map}, creating the generative agent population, and executing the main simulation loop. By organizing these processes into a modular structure, the \texttt{backend} ensures that the simulation can be easily adapted for various urban mobility scenarios.

\subsection{Transportation Network Management}

\begin{figure}
    \centering
    \includegraphics[width=0.9\linewidth]{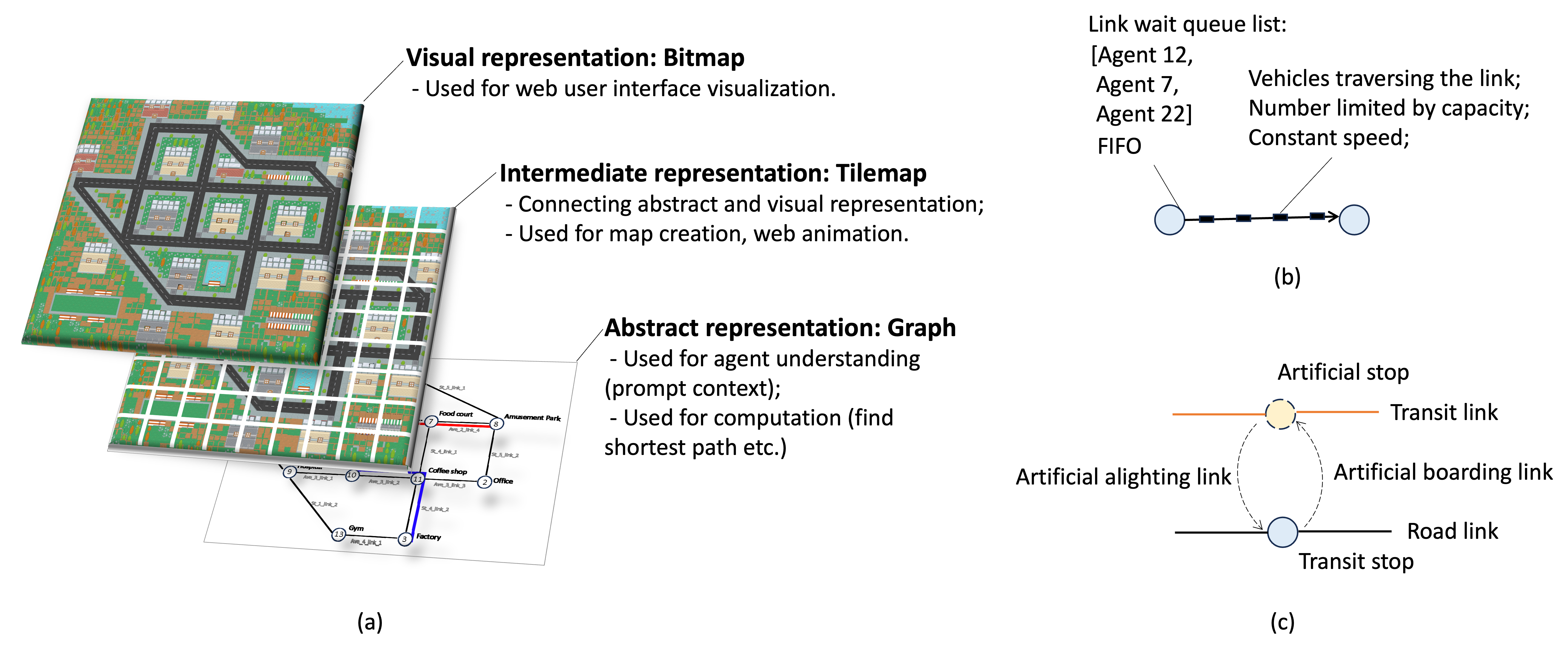}
    \caption{Transportation network representation in the simulation environment. The hierarchical system uses graph-structured data for LLM-based scenario generation, tilemaps for map editing and animation, and bitmap visualization for end-user interaction. Transit networks are modeled following Spiess and Florian's approach with artificial boarding and alighting links.}
    \label{fig:maze_representation}
\end{figure}

The simulation core employs a hierarchical transportation network representation strategy utilizing three complementary formats: graph, tilemap, and bitmap representations. Graph-structured data serves as the primary input for LLM-based scenario generation and reasoning. The transportation network is formally described by a graph $G=(V, E)$ where $V$ represents the set of vertices (nodes) and $E$ represents the set of edges (links). Activity locations such as apartments, offices, and schools are designated as ``facilities'', which exclusively serve as trip origins and destinations. Each facility is spatially associated with a specific node in the transportation network. Transit networks are modeled following the approach in \citet{spiess1989optimal}, representing transit lines as sequences of transit links that connect to the road network through artificial boarding and alighting links. This representation enables seamless multimodal routing while maintaining computational efficiency. For LLMs with image processing capabilities, GATSim supports bitmap input to enable visual reasoning about network topology and spatial relationships. Tilemap representations facilitate map editing operations and support real-time web interface animations, while bitmap-based visualizations provide the primary user interaction medium, as illustrated in Figure~\ref{fig:maze_representation}. 

Traffic dynamics are modeled using a point-queue system (a simplified traffic flow model where vehicles are treated as discrete entities that form queues at nodes and links, with instantaneous movement between queue positions rather than continuous spatial dynamics), where each network element maintains queues to govern agent movement. Link and facility capacity constraints are incorporated to simulate realistic congestion effects. Future versions will incorporate more sophisticated traffic flow models to enhance realism.

\subsection{Population Management}
\label{sec:population management}

At the start of the simulation, the simulation core either creates or loads the population, which consists of agent objects described in later sections. Each agent's activity state and location are updated continuously during the simulation loop. Agents' travel experiences are generated throughout the simulation and inserted into their respective memory modules. The simulation core may also provide constraint-checking functionalities. For instance, the simulation core enforces the requirement that each agent must return home by the end of the day. If this condition is not met, the agent is teleported home, and a reflection process is triggered, prompting the agent to consider why they failed to complete their journey as expected.

\subsection{User Interface} 

The user interface (Figure~\ref{fig:intro} (b)) of simulation environment serves four key functions: (1) rendering network topology and real-time agent states through a web-based visualization system; (2) accepting user input for simulation parameters such as run names, horizon, and configuration settings; (3) facilitating direct user-agent interaction, enabling users to interview agents to understand their decision-making processes and inject new ideas or constraints into agent cognition; and (4) supporting natural language scenario creation, allowing users to specify custom simulation conditions without programming expertise.

\section{Urban Mobility Foundation Model} 
\label{sec:llm}

The urban mobility foundation model refers to the LLM-based system that serves dual purposes: (1) generating synthetic population with realistic demographic profiles and behavioral characteristics, and (2) functioning as the cognitive engine that powers agent decision-making for mobility-related tasks including activity planning, travel decision making, and real-time adaptation.

\subsection{Population Synthesis}

GATSim generates synthetic population with carefully designed prompt that specifies demographic statistics including age distributions, household compositions, and socioeconomic characteristics (detailed in \texttt{root/create\_map} \citep{liu2025gatsim}). Recent research have shown that semi-structured, conversational representations yield more realistic agent behaviors than traditional parametric approaches \citep{park2024generative}. Following this insight, we adopt a narrative-based approach to represent agent characteristics, capturing not only basic demographics but also rich descriptive profiles that encompass lifestyle preferences, values, and behavioral tendencies (See Listing~\ref{list:example_persona_identity} for an example).

\begin{minipage}{\textwidth}
\begin{lstlisting}[
  basicstyle=\ttfamily\small,
  breaklines=true,
  caption={Example agent profile from synthesized population; a narrative-based approach is adpoted to represent agent characteristics, capturing not only basic demographics but also rich descriptive profiles that encompass lifestyle preferences, values, and behavioral tendencies.},
  label={list:example_persona_identity}
]
'name': 'Isabella Rodriguez'
'gender': 'female', 'age': 34,
'family_role': 'single', 'licensed_driver': True,
'work_facility': 'Coffee shop', 'occupation': 'barista',
'work_time': "7:00-16:00",
'preferences_in_transportation': 'prefer to travel alone and safer routes',
'innate': "friendly, outgoing, hospitable",
'lifestyle': "goes to bed around 11pm, wakes up around 6am",
'home_facility': 'Uptown apartment',
'household_income': 'medium', 'friends': ['Gloria Campbell', 'Michael Chen'],
'other_description': 'sabella is a cafe owner who loves making people feel welcome and creating a relaxing atmosphere. She likes discuss with Gloria Campbell on historical research and learns from Michael Chen about artifact preservation. '
\end{lstlisting}
\end{minipage}

Person profiles are pre-generated based on the synthesized population approach and then stored on disk. At the start of each simulation run, the simulation core loads these profiles and instantiates the corresponding agent objects as introduced in population management in Section~\ref{sec:population management}.

\subsection{Mobility Behavior Generation}

The generative processes within the urban mobility foundation model can be categorized into four types: planning, reactive behavior, chatting, and reflection. Model ensemble approach is adopted for efficiency (detailed in Table~\ref{tab:llm generations}). Complex and infrequent tasks such as planning warrant the use of large, relatively slower LLMs, whereas lightweight tasks such as chat summarization and importance scoring are better suited for smaller, more efficient models. The prompts used in the prototype are available at \citet{liu2025gatsim}. Each prompt contains variables representing the contextual information (``Vars in prompt'' column in Table~\ref{tab:llm generations}). These variables are dynamically generated by simulation environment.

\section{Agents} 
\label{sec:agents}

The generative agents represent the core innovation of the GATSim framework, embodying autonomous entities with cognitive architectures that enable realistic urban mobility behavior. The agent module integrates two key components: (1) a hierarchical memory that stores and retrieves relevant experiences (Sec~\ref{sec:memory}), and (2) planning, reactive, and reflection mechanisms that enable proactive schedule generation, dynamic adaptation to changing conditions, and transformation of specific travel experiences into generalized behavioral insights (Sec~\ref{sec:planning and reaction}). Correspondingly, in the prototype, the \texttt{Agent} module contains \texttt{memory\_modules} and \texttt{cognitive\_modules}.

\subsection{Agent Memory} 
\label{sec:memory}

The memory system represents a critical component of generative agents, enabling learning from experience, preference formation, and adaptive decision-making over time. The GATSim framework transcends simple information storage to create a cognitive architecture that mirrors human memory processes while addressing the unique requirements of urban mobility simulation. In the following sections, we first discuss memory structure and organization, then present memory retrieval and forgetting mechanisms.

\paragraph{Memory Structure and Organization}

The GATSim agent memory employs a hierarchical structure inspired by cognitive psychology, comprising short-term and long-term memory components with distinct but interconnected functions. Short-term memory serves as the agent's working cognitive space, maintaining dynamically updated information for the current simulation day including activity plans, ongoing states, travel paths, and immediate perceptions. Long-term memory stores accumulated information across extended time horizons, forming the foundation for behavioral learning and adaptation. 

To unify representation of diverse experience types—perceived events, social conversations, and internal reflections—we employ a ConceptNode class structure where each node encapsulates comprehensive attributes including textual content, semantic embeddings, extracted keywords, spatial coverage, temporal scope, importance scores, creation timestamps, and access histories. In contrast to previous implementations maintaining separate spatial memory structures \citep{park2023generative}, our unified approach integrates spatial-temporal information directly into long-term memory. This architectural decision enables cross-modal reasoning for adaptive behavior. For instance, an agent reflecting on metro delays can automatically connect congestion patterns with conversations about service disruptions or previously discovered alternative routes during similar situations. The unified structure simplifies retrieval mechanisms by enabling comprehensive memory search without requiring separate processing pipelines for different memory types.

\paragraph{Multi-Modal Memory Retrieval}

Effective retrieval mechanisms are fundamental to memory system functionality, enabling agents to access relevant past experiences for informed decision-making. Traditional memory retrieval systems typically rely on single-mode approaches such as keyword matching or semantic similarity, which often fail to capture the inherently spatial-temporal nature of urban mobility decisions. GATSim agent memory employs an multi-modal retrieval framework combining three complementary strategies: keyword matching, semantic similarity, and spatial-temporal relevance (Eqn~\ref{eqn:score}).

\begin{equation}
\label{eqn:score}
\text{score}_r(m) = w_r \cdot \text{matching}_r(m) + \delta \cdot \text{recency}(m) + \gamma \cdot \text{importance}(m)
\end{equation}
where $r \in \{\text{keyword}, \text{semantic}, \text{spatiotemporal}\}$; $w_r$ is the mode-specific weight; $\delta$ and $\gamma$ control the influence of importance and recency respectively, and importance is an LLM-assigned score reflecting the memory's significance.

Keyword matching score employs set-theoretic matching using the \textit{Jaccard similarity coefficient} \citep{real2001probabilisticjaccard} to identify memories with relevant keywords, supporting precise recall of specific events or locations (Eqn~\ref{eqn:matching_keyword}). The Jaccard coefficient balances precision and recall by considering both intersection and union of keyword sets.

\begin{equation}
\label{eqn:matching_keyword}
\text{matching}_{\text{keyword}}(K_q, K_m) = \frac{|K_q \cap K_m|}{|K_q \cup K_m|}
\end{equation}
where $K_q$ and $K_m$ represent the keyword sets of the query ($q$) and memory ($m$) respectively. This approach penalizes memories with excessive irrelevant keywords while rewarding those with focused, relevant content.

Semantic similarity leverages embedding-based search using FAISS \citep{douze2025faiss} for efficient vector-based operations (Eqn~\ref{eqn:matching_semantic}). Content embeddings are generated using pre-trained language models (\texttt{SentenceTransformer} with the model \texttt{all-MiniLM-L6-v2} \citep{wang2020minilm}), with semantic similarity computed as normalized cosine distance.

\begin{equation}
\label{eqn:matching_semantic}
\text{matching}_{\text{semantic}}(q, m) = 1.0 - \frac{d(e_q, e_m)}{d_{\max}}
\end{equation}
where $e_q$ and $e_m$ are the embeddings of the query and memory content respectively, enabling agents to generalize from past experiences through conceptual relationships.

Spatial-temporal relevance addresses the inherently spatial-temporal nature of urban mobility decisions through specialized mechanisms. The prototype adopts the \textit{Overlap Coefficient (Szymkiewicz–Simpson coefficient)} \citep{mcgill1979evaluation} for its desirable property that subset relationships yield perfect scores (Eqn~\ref{eqn:matching_overlap}, \ref{eqn:matching_spatialtemporal}).

\begin{equation}
\label{eqn:matching_overlap}
\text{overlap}(A, B) = \frac{|A \cap B|}{\min(|A|, |B|)}
\end{equation}
where $A$ and $B$ are two sets. This metric applies separately to spatial and temporal dimensions, computing spatial relevance for memories with spatial coverages $\mathcal{S}_q$ and $\mathcal{S}_m$ and temporal relevance for time intervals $\mathcal{T}_q$ and $\mathcal{T}_m$. The combined spatial-temporal relevance multiplies these scores, ensuring memories relevant to both location and timing receive priority.

\begin{equation}
\label{eqn:matching_spatialtemporal}
\text{matching}_{\text{spatiotemporal}}(q, m) = \text{overlap}(\mathcal{S}_q, \mathcal{S}_m) \times \text{overlap}(\mathcal{T}_q, \mathcal{T}_m)
\end{equation}

The prototype memory implementation also incorporates recency scoring through exponential decay modeling, recognizing that recent experiences often carry greater relevance for immediate decisions:

\begin{equation}
\text{recency}(m) = \lambda^{t_m}
\end{equation}

where $t_m$ represents time elapsed since memory creation or access, and $\lambda \in (0, 1)$ controls decay rate. The prototype defaults to exponential decay as it naturally models the rapid fading of routine events while preserving significant experiences, with recommended values of 0.90 for recent pattern emphasis or 0.95 for medium-term habit preservation. Table~\ref{tab:recency_scores} illustrates the practical implications of these parameters.

\begin{table}[h]
\centering
\caption{Recency scores under different decay rates.}
\label{tab:recency_scores}
\begin{tabular}{lcc}
\hline
\textbf{Time Elapsed} & \textbf{$\lambda = 0.90$} & \textbf{$\lambda = 0.95$} \\
\hline
1 day & 0.90 & 0.95 \\
1 week (7 days) & 0.48 & 0.70 \\
1 month (30 days) & 0.04 & 0.21 \\
\hline
\end{tabular}
\end{table}

\paragraph{Psychologically-Informed Forgetting Mechanisms}

Realistic memory decay modeling prevents unbounded memory growth while reflecting psychological principles governing human memory retention. Each memory entry receives an expiration attribute determining its lifespan based on non-linear relationships with importance scores, reflecting the well-established principle that salient experiences persist longer than mundane ones \citep{anderson1991reflections, wixted1991form}. The importance score criteria is shared at \citet{liu2025gatsim}. The lifespan calculation employs power functions aligning with empirical findings on human memory retention. See Appendix C for more details.

\subsection{Agent Activity Planning, Reaction and Reflection} 
\label{sec:planning and reaction}

The activity planning and reactive behavior represent core capabilities that distinguish generative agents from traditional rule-based approaches. This section presents the cognitive architecture underlying these capabilities, beginning with the activity planning process that synthesizes contextual information and social coordination to generate daily schedules. We then examine the dynamic plan revision mechanisms that enable real-time adaptation to changing environment. Afterwards, we discuss the multi-scale reflection mechanism that transforms specific travel experiences into generalized behavioral insights, enabling continuous learning and behavioral evolution over extended simulation periods. Finally, we present refinements that enhance the framework's robustness and performance.

\paragraph{Activity Planning Process}

\begin{figure}
    \centering
    \includegraphics[width=0.95\linewidth]{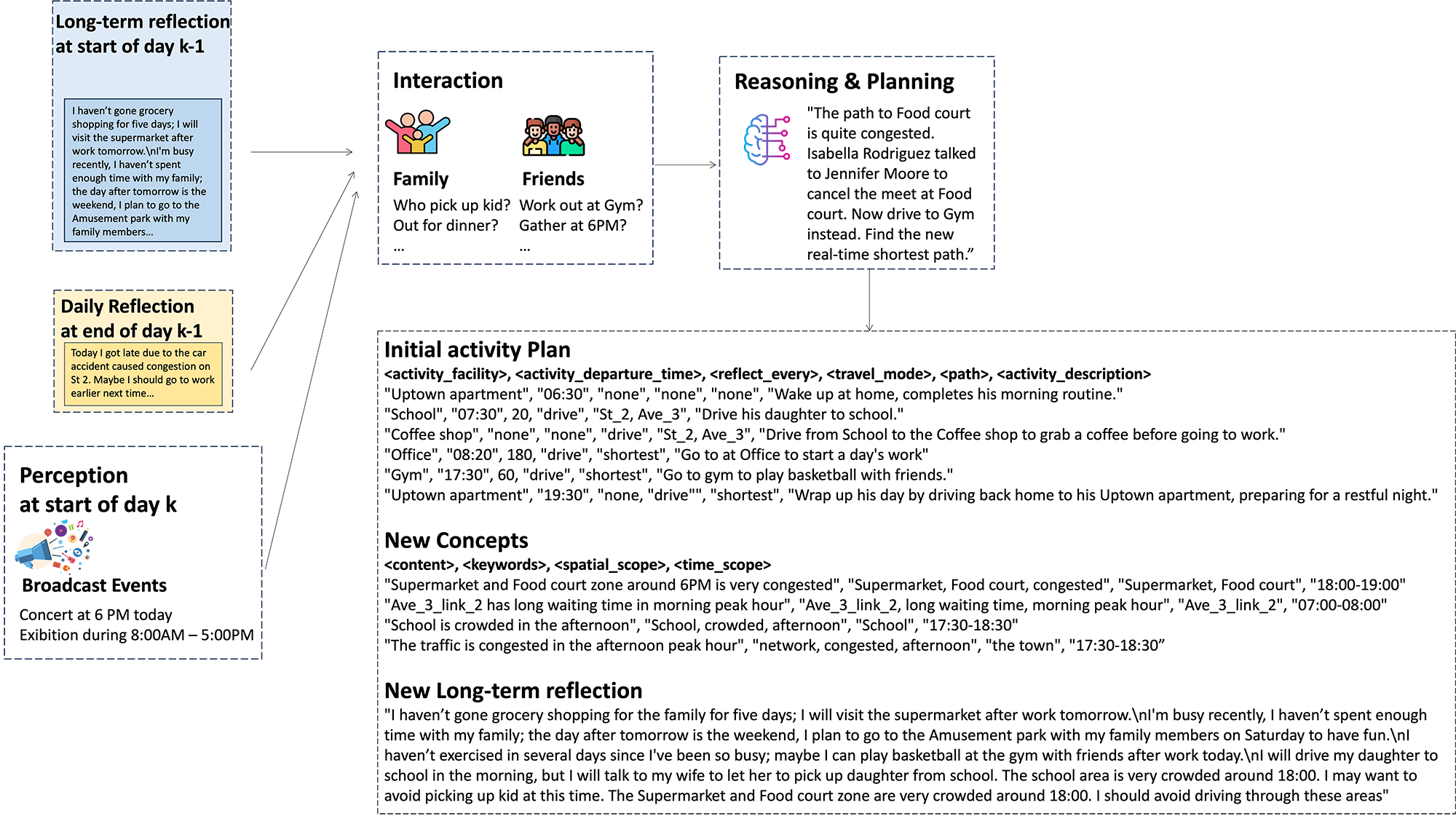}
    \caption{Generative agent activity planning process. Agents synthesize multiple information sources including the previous day's long-term reflection, initial activity plan, and daily reflection, combined with current broadcast events, retrieved relevant memories, and recent conversation summaries to generate today's updated long-term reflection,  comprehensive activity plan and concepts to keep in mind.}
    \label{fig:activity_plan_process}
\end{figure}

The activity planning process operates through a cognitive architecture that integrates multiple information sources across three stages (illustrated in Figure~\ref{fig:activity_plan_process}).

\textbf{Contextual Inputs Preparation Stage.} Each generative agent initiates daily planning by synthesizing contextual inputs from three primary sources: (1) long-term reflections that encapsulate key personal experiences spanning work-related tasks, social interactions, and family affairs while incorporating anticipations of upcoming events; (2) daily reflections from recent experiences that provide immediate behavioral context; and (3) perceived broadcast events representing shared external occurrences such as museum exhibitions, weather alerts, or infrastructure disruptions.

\textbf{Social Coordination Stage.} The planning process incorporates social coordination through collaborative decision-making where agents may initiate conversations with relevant parties before finalizing plans. For instance, household members might discuss school pickup responsibilities or friends might coordinate post-work activities. Conversation summaries are added to the memories of both parties, ensuring social agreements influence final plan generation.

\textbf{Reasoning and Planning Stage.} The final stage employs LLM-based processing to synthesize all contextual inputs and generate comprehensive activity plans. This approach enables agents to consider complex trade-offs between competing objectives, temporal constraints, social obligations, and personal preferences in ways that traditional optimization-based methods cannot capture. The LLM-based planner handles novel situations and unexpected combinations of constraints without requiring pre-programmed rules for every possible scenario.

\paragraph{Dynamic Plan Revision and Adaptation}

The revision process operates through carefully designed trigger mechanisms that balance responsiveness with efficiency, recognizing that excessive plan revision would be both computationally expensive and behaviorally unrealistic. Plan revision opportunities occur at strategically selected moments including: waiting periods at network nodes such as traffic intersections or transit stops; periodic intervals during ongoing activities based on reflection frequencies; and natural transition points when activities conclude.  

\begin{figure}
    \centering
    \includegraphics[width=0.95\linewidth]{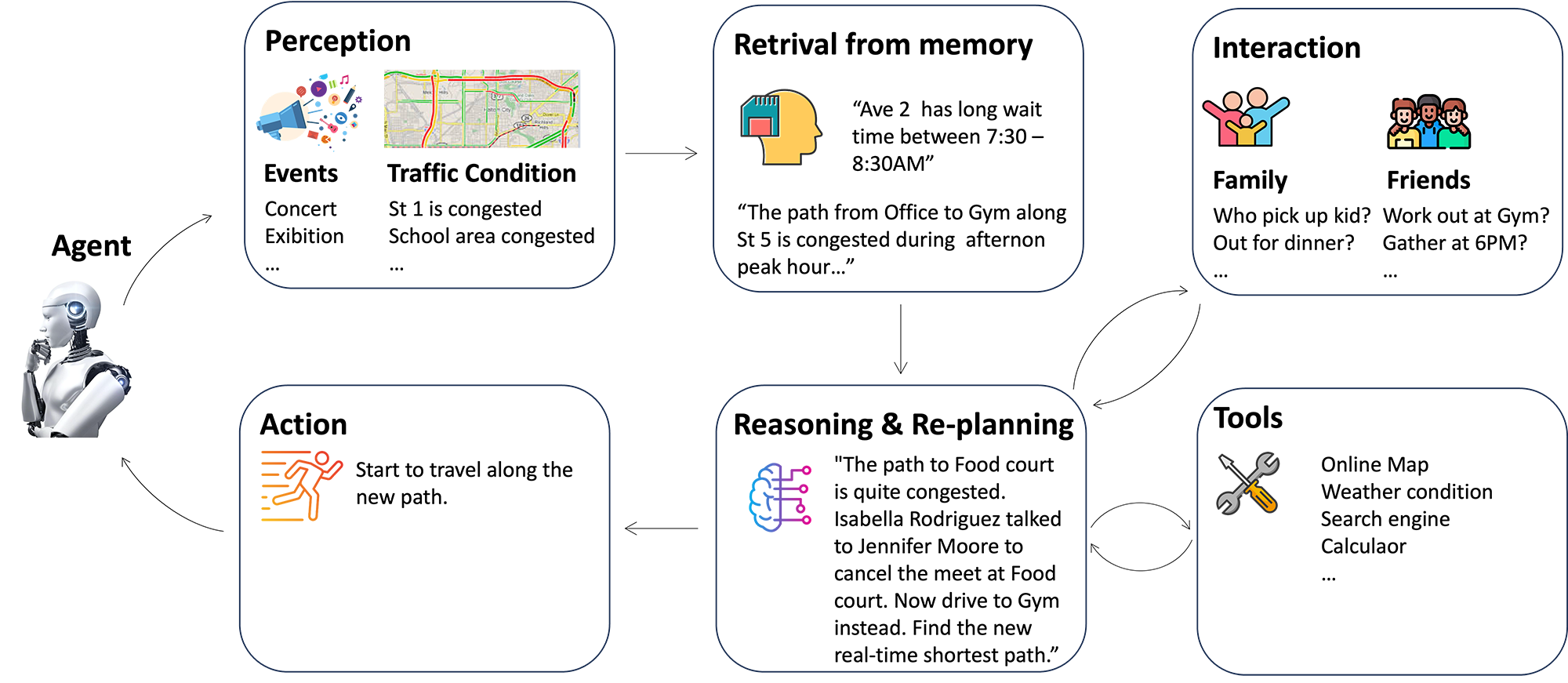}
    \caption{The agent action generation process, showing the flow from perception through reasoning to action execution. Plan revision occurs at strategically selected moments through four sequential stages: perception, memory retrieval, interaction, and reasoning with decision-making.}
    \label{fig:agent_action_process}
\end{figure}

The revision process operates through a four-stage cognitive pipeline that mirrors human decision-making processes while leveraging the reasoning capabilities of LLMs (Figure~\ref{fig:agent_action_process}): perception, memory retrieval, interaction, and decision-making. During the perception stage, real-time traffic state can be perceived by agents. Memory retrieval stage will retrieve today initial plan, all revision history, and memories related to current events as context. Agents can re-negotiate during interaction stage. The decision making output response ranging from plan continuation when no changes are needed to comprehensive plan replacement when major disruptions occur. Intermediate options include path updates to reflect real-time routing preferences, departure time adjustments to accommodate changed conditions, and partial plan modifications to address specific constraints while maintaining overall daily objectives. 

\paragraph{Multi-Scale Reflection and Learning}

\begin{figure}
    \centering
    \includegraphics[width=0.95\linewidth]{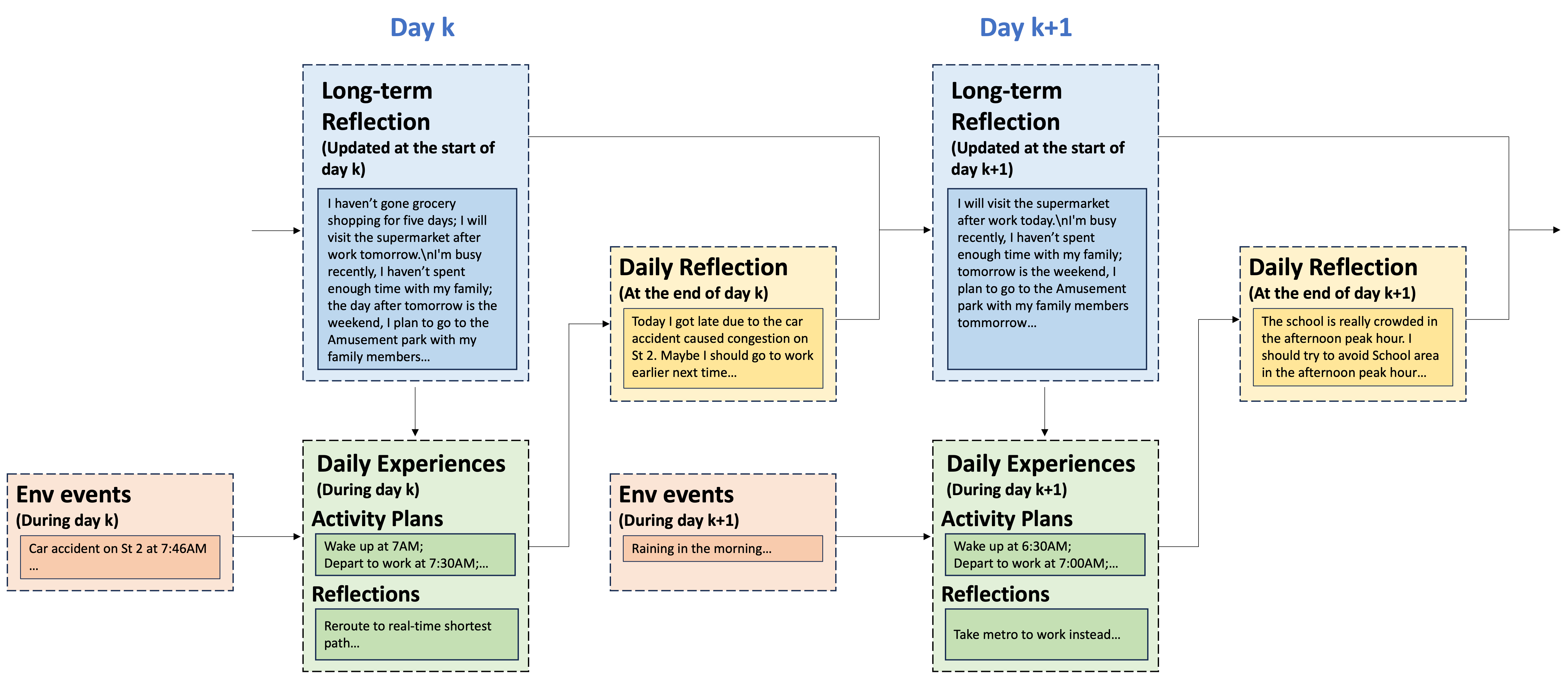}
    \caption{Generative agent reflection process. Agents operate multi-scale reflection mechanisms encompassing immediate reflection during reactive decision-making, daily reflection for experience synthesis, and long-term reflection for behavioral pattern consolidation.}
    \label{fig:reflection}
\end{figure}

The reflection mechanism is structured across multiple temporal scales to support both immediate adaptation and long-term learning. This multi-scale approach mirrors human cognitive processes where immediate experiences inform short-term decisions while accumulated patterns shape long-term behavioral strategies (Figure~\ref{fig:reflection}). 

\textbf{Immediate reflection} occurs during reactive decision-making. This reflection process creates concept nodes that capture the reasoning behind specific decisions, building a detailed history of decision-making contexts and outcomes that support future learning.

\textbf{Daily reflection} processes synthesize experiences from completed simulation days, identifying patterns, successful strategies, and areas for improvement. These reflections create higher-level insights that transcend specific incidents, enabling agents to develop generalized knowledge about traffic patterns, route reliability, and mode effectiveness that inform future planning decisions.

\textbf{Long-term reflection} maintains evolving summaries of behavioral patterns and preferences that persist across multiple simulation days. This reflection level enables the development of stable behavioral tendencies while maintaining flexibility for adaptation to changing conditions. The long-term reflection mechanism can potentially serve functionally as a day-to-day learning process that enables convergence toward traffic equilibrium states.

GATSim activity plan and revision mechanisms are able to ensure logical consistency and continuity of activity chains both within and across simulation days (illustrated by Listing~\ref{list:example_cross_day_continuity}). For intra-day consistency, each plan revision includes the full context of the today initial plan and revision history, ensuring agents maintain awareness of original intentions while adapting to dynamic conditions. For inter-day continuity, the hierarchical reflection mechanism creates temporal linkages through long-term reflections that track ongoing commitments, unfinished tasks, and multi-day plans. This recursive update process ensures that important activities postponed on one day (e.g., due to traffic disruptions) are appropriately rescheduled in subsequent days, creating realistic behavioral persistence across the simulation horizon.

\subsection{Model Refinements}

\paragraph{Limitation Analysis}
Detailed examination of agent responses revealed six categories of systematic errors that highlight current limitations and areas for improvement:
\begin{itemize}
    \item \textbf{Hallucination Problem}: Generative agents may generate reflections on non-existent traffic conditions. As shown in Listing~\ref{list:problem_hallucination}, an agent reflects that ``traffic on Ave\_3 has been consistently congested during the morning commute'' on the first day of simulation, despite having no prior experience with this route.
    
    \item \textbf{Instruction Following Problem}: Generative agents may fail to follow instructions properly. For example, in Listing~\ref{list:problem_instruction_following}, an agent repeatedly chatted about the same event with the same person despite our explicit instructions to avoid repetitive conversations.
    
    \item \textbf{Commonsense Violations}: Generative agents may generate actions that violate basic commonsense, such as planning to eat lunch at 19:00 or scheduling activities for time periods that have already passed (see Listing~\ref{list:problem_commensense}).
    
    \item \textbf{Reasoning Deficits}: Generative agents sometimes fail to make reasonable judgments and decisions. Examples in Listing~\ref{list:problem_reasoning} include an agent attempting to avoid congestion unrelated to planned route, another agent allocating insufficient time for a round trip to the Museum during lunch break, and frequent generation of invalid paths. This suggests limited spatial understanding capabilities in the underlying LLM.
    
    \item \textbf{Multi-agent Coordination Failures}: Multi-agent coordination refers to scenarios where multiple agents collaborate to accomplish shared tasks. The primary example in our experiments involves household coordination for childcare responsibilities. Other potential coordination scenarios include friends arranging to meet for social activities and ride-sharing. However, coordination among agents frequently failed. During the communication stage, agents sometimes failed to achieve clear task allocation, leading to ambiguous agreements, incomplete coverage, mutual deferral, or duplication of effort (see Listing~\ref{list:problem_multiagent_coordination}). Even when agents initially reached consistent plans, subsequent dynamic plan revisions during the day could break the coordination without re-negotiation.
    
    \item \textbf{Over-responsiveness and Submissiveness Behavior}: In our experiments, generative agents are often overly responsive to network events (such as exhibitions or entertainment events), despite being assigned different personality traits and preferences. Additionally, generative agents tend to be excessively accommodating, barely rejecting proposals from family members or friends. This behavior likely stems from the fact that most LLMs are pre-trained to be helpful assistants, and this characteristic persists even when instructed to adopt different personas.
\end{itemize}

While these limitations stem from fundamental challenges in contemporary LLM architectures and training paradigms rather than inherent framework deficiencies, they nonetheless indicate promising avenues for future model development and specialized fine-tuning methodologies. Despite these constraints, we propose several practical interventions below that can substantially alleviate these issues within current technological bounds.

\paragraph{Proposed Remedies}
To address the identified limitations, we propose a comprehensive set of practical strategies that can be implemented within current technological constraints:
\begin{itemize}
    \item \textbf{Leveraging Advanced Base Models}: Employing LLMs with enhanced reasoning capabilities represents the most direct and convenient approach to improving generative agent performance. We experimented with different versions of Qwen models and found that agents using the \texttt{qwen\_max} generate more reasonable activity plans and exhibit significantly better coordinated behavior compared to the smaller \texttt{qwen\_plus} model, albeit at the cost of increased latency. Concurrent API calls and batched processing can effectively mitigate latency concerns while preserving quality improvements.
    
    \item \textbf{Iterative Activity Plan Validation}: Implementing systematic plan review processes enables agents to self-correct errors through multiple validation rounds. For example, an agent initially generated a plan to use transit for the entire day except for one midday activity; after multiple checks, the agent successfully identified and corrected this inconsistency. Generative agents may occasionally forget to include the ``return home'' activity at the end of their plan, but they typically remedy this oversight before day's end. Since agents naturally adapt plans throughout the day, perfect initial planning is unnecessary — continuous refinement proves more effective and realistic.
    
    \item \textbf{Task Decomposition}: Generative agents using smaller LLMs (such as 7B parameter models) struggle to simultaneously generate reflections and daily activity plans with consistent vehicle usage and task coordination in a single step. When practical constraints necessitate using smaller LLMs, we recommend decomposing the task into smaller components, such as reflection, activity facility decision, travel mode decision, and departure time decision etc. Even for agents with larger LLMs, task decomposition remains beneficial, as demonstrated by the success of Chain-of-Thought reasoning \citep{wei2022chain}.
    
    \item \textbf{Prompt Optimization}: Long contextual inputs can overwhelm agent attention mechanisms, degrading instruction adherence despite expanded context windows in modern LLMs. Prompt engineering techniques can improve generative agent performance, such as placing important and frequently ``forgotten'' instructions at the end of prompts and incorporating redundancy for critical instructions.
    
    \item \textbf{Fine-tuning for Behavioral Alignment}: Hallucination presents a dual challenge — enabling creative behavioral diversity while maintaining behavior grounded in actual experiences. Fine-tuning offers a promising approach to reducing unwanted hallucination while simultaneously enhancing behavioral diversity.
    
    \item \textbf{Rule-based Validation}: Implementing automated validation for critical domain constraints (e.g. vehicle usage consistency) provides immediate error detection without requiring model modifications. When violations occur, agents receive specific feedback prompting plan regeneration. This lightweight approach offers immediate practical benefits with minimal implementation overhead.
    
    \item \textbf{Appropriate Abstraction Level Selection}: Effective simulation requires strategic decisions about modeling granularity based on research objectives and computational resources. Key considerations include: vehicle sharing mechanisms within households, agent perception radius and frequency, social network complexity and evolution, and temporal resolution for activity scheduling. These choices fundamentally shape both computational requirements and behavioral realism, requiring careful calibration for specific application domains.
\end{itemize}

\section{Experiments} 
\label{sec:experiments}

To validate the GATSim framework, we conduct experiments at both microscopic and macroscopic levels. These experiments serve as concept validation for the GATSim framework. Despite the limited simulation size, the key contributions of this study—including the memory system, activity-planning pipeline, and reactive / reflective mechanisms—can be effectively examined and validated through small-scale scenarios. Such examples can demonstrate how these components interact to produce coherent decision-making and adaptive behaviors, which form the conceptual foundation of large-scale applications. The complete simulation configuration and results are available at \citet{liu2025gatsim}.

\paragraph{Urban Mobility Foundation Model}
To facilitate rapid prototyping and framework validation, we adopt off-the-shelf LLMs without domain-specific fine-tuning. The prototype utilizes \texttt{GPT-4o} \citep{achiam2023gpt} for population synthesis due to its strong capability in generating coherent long-form text. \texttt{Qwen} model family \citep{yang2024qwen2} is employed for agent behavioral modeling for their favorable balance between performance and cost efficiency. They represent leading closed-source and open-source models, respectively, at the time of writing. The Qwen model family offers multiple variants of varying sizes: the largest model, \texttt{qwen-max-2025-01-25}, is employed for activity planning; the medium-sized \texttt{qwen-plus-2025-07-14} is used for reactive behavior and reflection generation; and the smallest but most efficient variant, \texttt{qwen-turbo-2025-07-15}, is applied to remaining lightweight tasks. In our experiments, each generative agent required around one million inference tokens for each simulation day.

\begin{figure}
    \centering
    \includegraphics[width=0.75\linewidth]{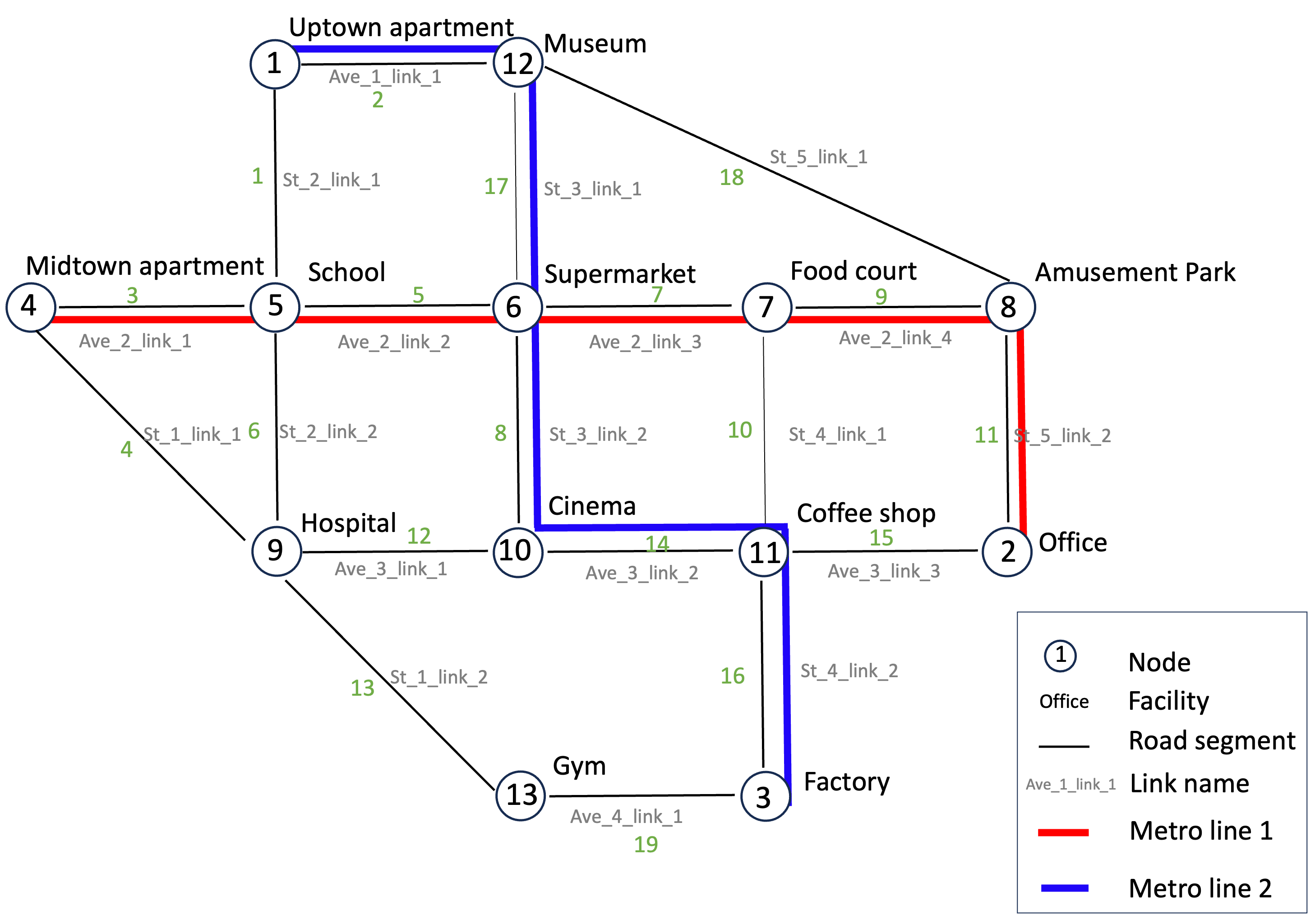}
    \caption{Transportation network topology used in experiments, adapted from the Nguyen-Dupuis network with added metro lines.}
    \label{fig:network_graph}
\end{figure}

\paragraph{Transportation Network}
The prototype is tested on a stylized network adapted from the Nguyen-Dupuis network \citep{nguyen1984efficient}. The network consists of 13 nodes connected by road links and two metro lines (\texttt{Metro\_1} and \texttt{Metro\_2}), creating a multimodal transportation environment. In all experiments, agents could choose between two travel modes: solo driving (private car) and walk + public transit (walking on road links combined with metro use). While our current experiments focus on these two modes, GATSim framework is designed to accommodate additional travel modes. The agent cognitive architecture can naturally reason about more complex mode choices (including carpooling coordination etc.) without fundamental architectural changes. Each node hosts one facility, creating 13 distinct activity locations. This configuration represents a simplified but realistic urban structure. The complete transportation network topology is illustrated in Figure~\ref{fig:network_graph}. In this small-scale example, congestion is induced by enforcing strict link capacity constraints (each road link allows a maximum of two vehicles at a time), enabling agents to experience realistic congestion delays and journey times.

\paragraph{Synthetic Population Generation}
A synthetic population of 70 individuals is generated using GPT-4o with carefully designed demographic constraints to ensure realistic population structure (detailed in \texttt{root/create\_map} \citep{liu2025gatsim}). Practically, generating an entire city-scale population through a single LLM query is not feasible. Beyond our implementation, an emerging line of research has explored large-scale population synthesis using deep generative models~\citep{kim2023deep, ge2024scaling}. The demographic composition follows realistic urban patterns (summarized in \ref{tab:population_summary}. Family sizes range from 1 to 4 members, and social networks include 0 to 3 friends per individual. Employment is concentrated in two major centers: Factory (20 workers) and Office (17 workers), representing typical industrial and service sector employment. Other facilities employ 1-3 staff members each, creating realistic employment distribution.

\paragraph{Simulation Constraints and Simplifications}
Several simplifications are implemented in experiments. Travel mode is determined at the beginning of each simulation day and remains fixed throughout. Children's activities are implicitly represented through parental coordination behaviors such as school drop-off and pick-up activities. These simplifications are adopted solely for experimental purposes and do not reflect inherent limitations of GATSim.

\subsection{Experiment 1: Microscopic Behavior Validation}

This section evaluates the realism of individual agent decision-making by comparing generative agent behaviors with human responses in controlled mobility scenarios. The evaluation focuses on two key capabilities: reactive decision-making in dynamic conditions and behavioral learning through experience accumulation. We conducted comparison between generative agents and human annotators using role-playing. Five human participants were recruited as annotators, each presented with 10 mobility scenarios carefully selected from the simulation for a total of 50 evaluation cases. Each scenario included comprehensive contextual information: simulation environment description, agent role profile, real-time traffic conditions, environmental perceptions, retrieved memory content. Both generative agents and human annotators were asked to provide two outputs for each scenario: a reflection capturing their reasoning process and an action plan for immediate next steps. To ensure unbiased evaluation, we employed GPT-o1 as an independent judge \citep{zheng2023judging}, randomly permuting the order of AI and human responses to eliminate positional bias effects. The LLM judge outputs which response is more realistic, or a ``tie''. The evaluation of ``being realistic'' encompasses multiple dimensions, including contextual appropriateness (alignment with the given scenario, agent profile, and environmental context), reasoning coherence (logical consistency and justification of the decision process), behavioral plausibility, and human-like priorities \citep{liu2025gatsim}.

Note that we are comparing generative agents with human annotators under role-playing tasks rather than comparing with humans with their actual profiles and observed real-world behaviors. This choice reflects practical constraints in transportation research. Recruiting a large and diverse populations to participate in extended mobility simulations using their actual profiles is challenging and raises privacy concerns. Role-playing by trained annotators provides a controlled, replicable benchmark that captures human reasoning patterns while maintaining experimental validity.

\subsubsection{Behavioral Capability Results}

Prior to quantitative comparison, we observed that generative agents exhibit sophisticated behavioral capabilities across multiple dimensions:

\textbf{Dynamic Adaptability}: Agents demonstrate flexible, context-aware decision-making in response to traffic disruptions and unexpected events. Examples include real-time route adjustment when encountering congestion, departure time modifications based on perceived delays, and mode choice reconsideration when service disruptions occur (see Listing~\ref{list:example_reactive_behaviors_short}, Listing~\ref{list:example_reactive_behaviors} for examples).

\begin{lstlisting}[
  basicstyle=\ttfamily\footnotesize,   % smaller font
  lineskip=-1pt,                       % reduce line spacing
  aboveskip=4pt, belowskip=4pt,        % reduce vertical gaps before/after
  breaklines=true,
  escapeinside={(*}{*)},
  caption={One example of generative agent flexible reactive behavior.},
  label={list:example_reactive_behaviors_short}
]
Name: George Wright | Time: Thu 2025-03-13 07:28
 I experienced a 9-minute wait at Metro_1_link_1 this morning, which was unexpected and slightly delayed my arrival at the Office. While there is currently only light congestion on Ave_2_link_2 and moderate congestion on Ave_1_link_1, the earlier-than-expected delays on Metro_1 suggest potential variability in transit reliability today. This could affect my ability to comfortably return from the Museum visit during lunch if similar conditions occur on my outbound trip back to Office. However, since no major disruptions are reported and my travel mode remains consistent, I will maintain the current plan with minor timing adjustments.
Plan revisions: George Wright update the departure time of next activity at Museum to 12:45
\end{lstlisting}

\textbf{Experience-Based Learning}: Agents successfully acquire knowledge through reflective processes, transforming specific travel experiences into generalizable insights. This learning manifests as improved route selection based on historical congestion patterns, refined departure time strategies, and evolved preferences for reliable travel options (see Listing~\ref{list:example_reflections_short}, Listing~\ref{list:example_reflections} for examples).

\begin{lstlisting}[
  basicstyle=\ttfamily\footnotesize,   % smaller font
  lineskip=-1pt,                       % reduce line spacing
  aboveskip=4pt, belowskip=4pt,        % reduce vertical gaps before/after
  breaklines=true,
  escapeinside={(*}{*)},
  caption={One example of generative agent gaining knowledge through reflections on travel experiences.},
  label={list:example_reflections_short}
]
Rebecca Lee | Date: Thu 2025-03-13 AM
My early morning workout routine at the Gym has proven effective not only for maintaining fitness but also for avoiding peak traffic, particularly on St_1_link_2, which tends to be congested during morning hours. Yesterday's adjustment to an alternative route via St_4_link_2 and Ave_3_link_3 was successful, highlighting how critical it is to stay aware of recurring congestion patterns and have contingency plans in place.
\end{lstlisting}

\subsubsection{Comparative Performance Results}

Of the 50 scenarios evaluated, generative agents were judged more realistic in 23 cases (46.0\%), human annotators prevailed in 20 cases (40.0\%), and 7 cases (14.0\%) resulted in ties. The slight advantage for generative agents can be attributed to their superior ability to process and synthesize extensive contextual information. Human annotators frequently struggled with the volume of information presented, including detailed activity histories, event logs, social interaction records, and environmental conditions. This result reflects the difference in effective ``context windows'' between LLMs and human cognition. While LLMs can maintain attention across extensive textual contexts, humans have more limited working memory but compensate with stronger spatial reasoning and common-sense grounding. We emphasize that our claim is not that LLM judge are ``better'' than humans in absolute terms, but rather that they demonstrate competitive behavioral realism in role-playing scenarios while offering advantages in information processing capacity and consistency.

\begin{figure}
    \centering
    \includegraphics[width=0.95\linewidth]{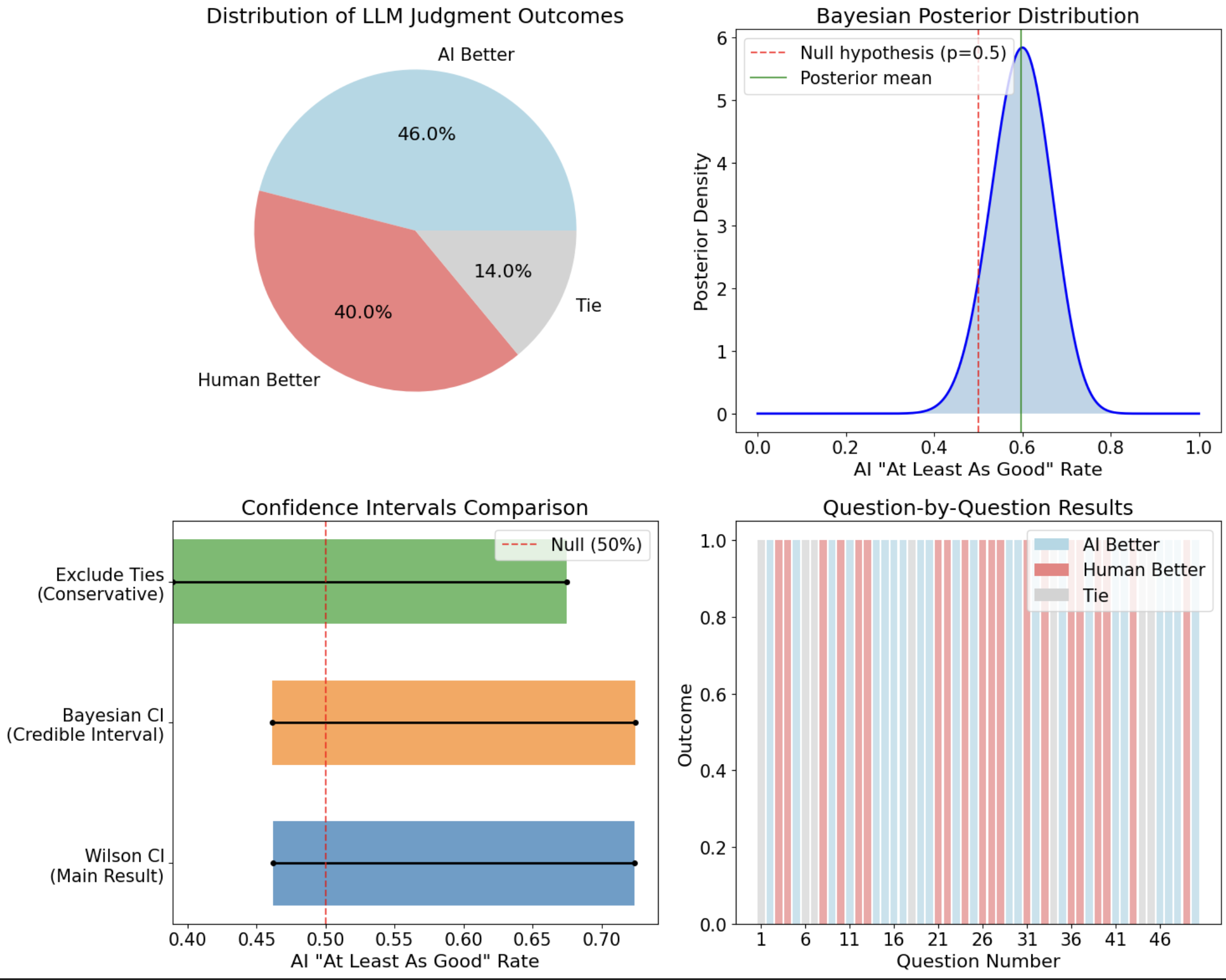}
    \caption{Statistical analysis of comparative performance between generative agents and human annotators, showing probability distributions and confidence intervals.}
    \label{fig:test}
\end{figure}

We conducted statistical analysis to validate the observed performance differences. First, we verified that the LLM judge produces meaningful (non-random) evaluations rather than arbitrary decisions ($p = 0.002$). Second, we tested whether generative agents significantly outperform human annotators. The statistical test shows no significant difference ($p = 0.380$), indicating competitive performance. Using Bayesian analysis, we find a 92\% probability that generative agents perform at least as well as human annotators, providing strong evidence for behavioral realism (Figure~\ref{fig:test}). In summary, GATSim agents demonstrate human-comparable behavioral realism in mobility decision-making scenarios.

\footnote{Technical details: We used a binomial proportion test to verify non-random judge performance ($H_0$: $p_{\text{win}} \leq 0.333$) and a sign test to compare agent and human performance ($H_0$: $p_{\text{AI}} \leq 0.5$).} 

\subsection{Experiment 2: Macroscopic Pattern Validation}

\begin{figure}
    \centering
    \includegraphics[width=0.98\linewidth]{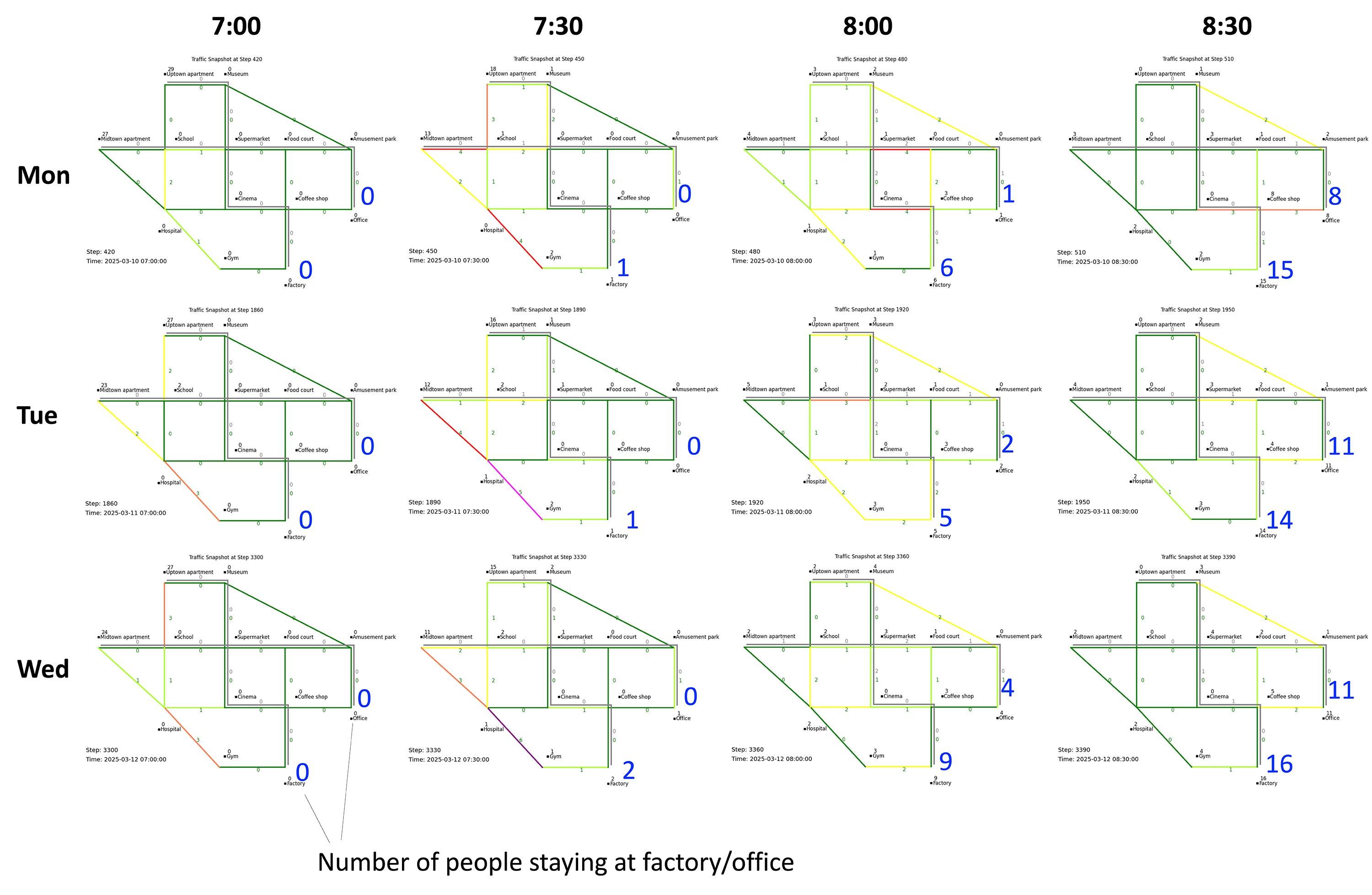}
    \caption{Traffic state evolution across three simulation days, showing snapshots at critical time points 7:00, 7:30, 8:00, and 8:30. The progression demonstrates emergent peak spreading and congestion adaptation as agents learn from travel experiences and adjust their departure strategies.}
    \label{fig:experiment_2}
\end{figure}

This experiment examines emergent system-level traffic patterns arising from individual agent learning and adaptation over multiple simulation days. The simulation spans three consecutive days beginning Monday, March 10, 2025. Employees at Factory and Office predominantly follow 8:00–17:00 schedules. This creates morning and evening peak periods, enabling observation of traffic pattern evolution. Two specialized visualization tools were developed to capture spatiotemporal traffic dynamics: ``traffic state animator'' for continuous temporal visualization and ``traffic state snapshot'' for discrete-time network state analysis. Figure~\ref{fig:experiment_2} presents traffic state snapshots at critical time points 7:00, 7:30, 8:00, and 8:30 across the three simulation days. 

\textbf{Observed Traffic Evolution Results} On Monday, most agents depart home around 7:20, allowing approximately 40 minutes to reach their workplace under free-flow conditions. However, the resulting congestion causes only 7 agents to arrive punctually at 8:00. As agents accumulate travel experiences and update their mental models through reflection, they adaptively adjust departure times, routes, and mode choices. The temporal evolution is evident in Figure \ref{fig:experiment_2}'s first column: the morning peak progressively shifts earlier from Monday to Wednesday. While the network remains uncongested at 7:00 on Monday, mild congestion emerges by Wednesday at the same hour. Concurrently, traffic disperses more broadly across the network, with a modest increase in transit ridership. The observed traffic evolution across the three simulation days demonstrates GATSim's equilibrium-seeking behavior through distributed agent learning. The progressive peak spreading and congestion reduction reflect agents' long-term reflections guiding them toward more efficient travel strategies. While the simulation period is insufficient to observe complete convergence, the trends suggest movement toward an equilibrium state through the reflection-based learning mechanism. Over the past several decades, transportation researchers have developed various day-to-day dynamics models (like Evolutionary Algorithms used by MATSim) to simulate how travelers adjust their choices over time to reach equilibrium states. GATSim's approach can be understood as a cognitively-grounded alternative to these classical methods.

\subsection{Experiment 3: Macroscopic Incident Response Validation}

This experiment examines how agents collectively adapt to infrastructure incidents through distributed learning and behavioral adjustment. The analysis focuses on both immediate reactive behaviors and longer-term adaptation patterns that emerge from agent experience accumulation. Phase 1 involves a week-long simulation from March 10-16, 2025, allowing agents to develop stable travel patterns and accumulate experience with normal network conditions. Phase 2 introduces a network incident during the second week to observe agent adaptation responses. The  incident introduced is capacity reduction on link segment ``Ave\_2\_link\_2'' dropping from 2 to 1 (50\% reduction) on Tuesday, March 18, spanning the morning peak period from 7:30 to 8:30.

\begin{figure}[H]
    \centering
    \includegraphics[width=0.8\linewidth]{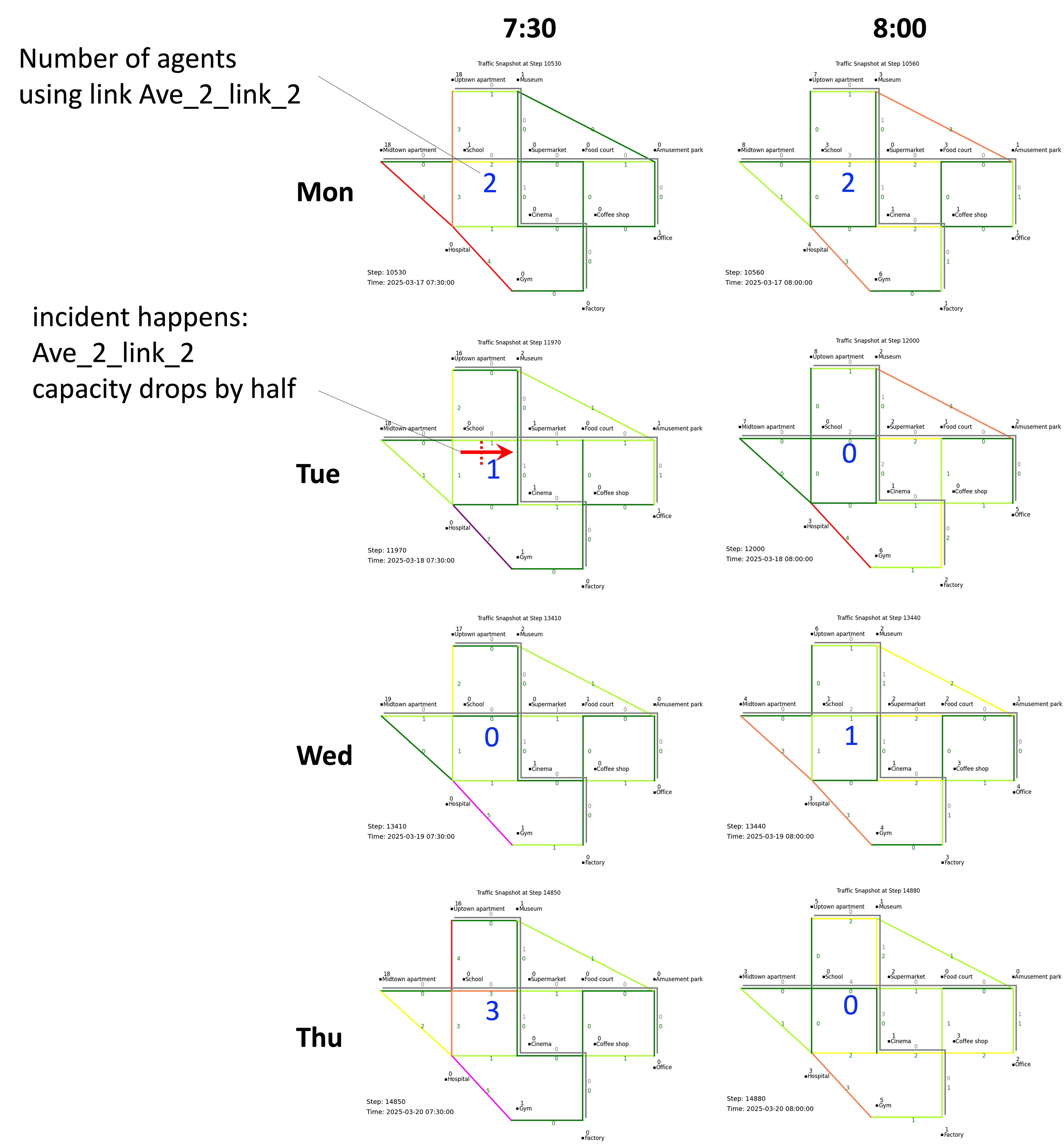}
    \caption{Network incident response evolution across four consecutive days. The first and second rows show traffic states at 7:30 and 8:00 respectively, demonstrating agent adaptation to a capacity reduction incident on Ave\_2\_link\_2 (Tuesday, March 18, 7:30-8:30). The progression illustrates immediate disruption effects, subsequent over-reactive avoidance behavior, and gradual recovery patterns as agents adapt their travel strategies based on accumulated experiences.}
    \label{fig:experiment_3}
\end{figure}

\paragraph{Observed Behavioral Adaptation Results} Figure~\ref{fig:experiment_3} presents traffic state snapshots at 7:30 and 8:00 across four consecutive days (Monday, March 17 through Thursday, March 20), illustrating the temporal evolution of agent responses to the network incident.

\textbf{Immediate Response (Tuesday, March 18):} Agents encounter unexpected congestion during the incident period, resulting in delayed arrivals and accumulated negative travel experiences. The capacity reduction creates a bottleneck that affects not only agents using the specific link but also generates spillover effects on parallel routes.

\textbf{Short-term Adaptation (Wednesday-Thursday, March 19-20):} Agents who experienced delays during the incident demonstrate proactive behavioral adjustments in subsequent days. Despite the incident's resolution, affected agents continue to modify their travel strategies based on accumulated experiences. Key adaptations include:

\begin{itemize}
\item \textbf{Temporal Shifting}: Agents who previously used Ave\_2\_link\_2 during peak hours systematically depart earlier to avoid perceived congestion risks, resulting in reduced traffic volumes at 8:00 compared to pre-incident levels.

\item \textbf{Route Avoidance}: The agent population even exhibits over-reactive behavior, with Ave\_2\_link\_2 flows dropping to near-zero levels during morning peak periods on Wednesday and Thursday. This demonstrates how negative experiences can lead to persistent avoidance behaviors even after disruption resolution.

\item \textbf{Gradual Recovery}: While traffic flows at certain time points (e.g., 8:00 on Thursday) may approach or temporarily exceed pre-disruption levels due to temporal concentration of demand, analysis of aggregate daily link usage reveals that total traffic volumes on Ave\_2\_link\_2 remained substantially below baseline Monday levels. This suggests that agent confidence in route reliability rebuilds gradually through positive experience accumulation.
\end{itemize}

\paragraph{System-Level Implications} The experiment reveals several important characteristics of generative agent responses to network disruptions:

\textbf{Distributed Information Processing}: Agents process incident experiences independently, leading to heterogeneous adaptation strategies that reflect realistic individual differences in risk tolerance and learning patterns.

\textbf{Collective Over-reaction}: The population demonstrates collective over-reaction to the incident, with persistent avoidance behaviors exceeding what would be optimal given the temporary nature of the disruption. This behavior mirrors observed phenomena in real traffic systems where incidents create lasting behavioral changes.

\textbf{Asymmetric Recovery}: The recovery process exhibits asymmetric patterns, with agents showing greater reluctance to return to previously disrupted routes compared to their willingness to initially abandon them. This hysteresis effect reflects realistic behavioral inertia in transportation choice.

These findings demonstrate that GATSim can capture realistic behavioral responses to network incidents, including both immediate adaptation and longer-term behavior. The framework's ability to model these complex behavioral dynamics without explicit programming of incident response rules highlights the potential for more realistic transportation system analysis under disruption scenarios.

\section{Conclusion} 
\label{sec:conclusion}

This work presents GATSim, a novel framework that introduces generative agents to transportation simulation, contributing to the fundamental paradigm shift from rule-based to intelligence-based modeling. The framework enables agents to continuously learn and adapt through dedicated memory systems and reflection mechanisms, creating realistic behavioral evolution over time. GATSim demonstrates the potential to mimic emergent properties of human societies, revealing underlying mechanisms of urban mobility that are difficult to capture through conventional modeling approaches. The primary contributions of this study are the comprehensive GATSim framework and the specialized agent cognitive architectures encompassing memory, planning, reaction, and reflection mechanisms tailored for urban mobility simulation.

The microscopic validation reveals that generative agents perform competitively with human annotators in role-playing scenarios. The macroscopic validation demonstrates realistic traffic evolution over multiple simulation days, with agents adaptively adjusting departure times, routes, and mode choices to achieve system states comparable to traditional user equilibrium models.

The GATSim framework opens numerous avenues for practical applications in transportation planning and policy analysis. It can enable planners to easily specify scenarios and analyze outcomes without extensive programming expertise. The behavioral richness of generative agents supports detailed analysis of policy interventions, infrastructure changes, and service disruptions with greater realism than traditional approaches. The framework's ability to model social coordination and information sharing makes it particularly valuable for studying ride-sharing, transit coordination, and crisis response scenarios. Furthermore, the adaptive learning capabilities enable longitudinal studies of behavioral change in response to new transportation services, pricing policies, or urban development patterns. The framework's modular design facilitates integration with existing transportation models and data sources, supporting practical deployment in real-world planning contexts.

\paragraph{Limitations} Despite promising results, several limitations constrain current framework capabilities. LLM agents demonstrate excessive cooperativeness compared to human behavior, reflecting their training as helpful assistants rather than realistic individual personas, potentially overestimating cooperation levels in transportation systems. Multi-modal urban environment data processing remains challenging, as LLMs still struggle with complex spatial reasoning despite our framework's spatial query mechanisms. Memory retrieval optimization and consistent behavior maintenance over extended simulation periods present ongoing challenges. The current traffic flow model employs a simplified point-queue approach without intersection signal control, turn movement modeling, or realistic capacity representations. Validation benchmarks for LLM-driven agent-based simulation remain an open problem, with current methods relying on role-playing scenarios rather than direct real-world behavior comparisons. Finally, computational intensity limits practical deployment to small populations, though advances in model efficiency may address these constraints over time.

\paragraph{Future Directions} Several promising research directions emerge from this work. Generic LLM reasoning may not capture transportation-specific decision heuristics that real travelers employ, such as ``always add 10 minutes to Google Maps estimates''. Incorporating these domain-specific heuristics requires careful prompt engineering or specialized training on transportation behavior datasets. The current implementation employs static, hand-crafted social structures including family relationships, friendship networks, and occupational roles that remain fixed throughout simulation. While generative agents in principle could model dynamic social structure evolution — allowing relationships to form, strengthen, or dissolve based on agent interactions — this extension would require substantially longer simulation horizons and introduces complexity beyond transportation-focused validation. Future work could explore and evaluate this extension for studies specifically examining the interplay between social dynamics and mobility patterns. Defining ``realistic'' behavior is fundamentally challenging and inherently subjective. We adopted the LLM-as-a-judge approach for its consistency, reproducibility, and scalability. Future research could pursue alternative validation frameworks, such as evaluating the degree to which generated behaviors remain ``in-domain'' by comparing their distributional properties with observed real-world travel data, or developing hybrid approaches that combine automated judgment with domain expert evaluation. Advances in model compression, specialized hardware, and LLM inference efficiency will be crucial for practical deployment at city scale. Finally, comprehensive validation requires three-way comparisons among real-world observations, traditional simulation baselines, and generative agent framework results at realistic population scales.

\section*{Acknowledgment}
This research was funded by the National Natural Science Foundation of China (grant number 52325210, 52131204, 52402407) and Shanghai Baiyulan Talent Project Pujiang Program (grant number 24PJD115).

\clearpage
\appendix

\section{Main Simulation Loop of GATSim Prototype}

The simulation core operates through discrete-time orchestration of agent behaviors and system dynamics, as shown by Algorithm~\ref{alg:main_loop}. Each timestep follows a structured sequence: (1) plan generation and revision for adaptive replanning, (2) activity state updates managing transitions between activities, (3) mobility execution within the transportation network, (4) selective experience reflection for behavioral learning, and (5) state persistence with time advancement. This cyclical process continues until reaching specified simulation horizons, typically 24 hours for daily studies or longer periods for behavioral adaptation analysis.

\refstepcounter{appendixaalgorithm}
\label{alg:main_loop}
\begin{algorithm}[H]
\textbf{Algorithm \ref{alg:main_loop}.} Main Simulation Loop of GATSim Prototype.
\begin{algorithmic}[1]
\While{$t < T_{end}$}
    \State \textbf{Step 1: Plan Generation/Revision and reflection}
    \For{each agent $a \in$ Agents}
        \If{NeedsPlanUpdate($a, t$)}
            \State $a$.GenerateOrRevisePlan()
        \EndIf
    \EndFor
    
    \State \textbf{Step 2: Activity State Updates}
    \For{each agent $a \in$ Agents}
        \If{ActivityComplete($a$) \textbf{or} DepartureTimeDue($a, t$)}
            \State $a$.UpdateActivityState()
        \ElsIf{EnRoute($a$) \textbf{and} NextActivityDue($a, t$)}
            \State $a$.InterruptCurrentTravel()
        \EndIf
    \EndFor
    
    \State \textbf{Step 3: Network Movement}
    \State ExecuteNetworkMovement($t$)
    
    \State \textbf{Step 4: Experience Reflection}
    \For{each agent $a \in$ Agents}
        \If{ShouldReflect($a, t$)}
            \State $a$.DailyReflection()
        \EndIf
    \EndFor
    
    \State \textbf{Step 5: State Persistence}
    \State SaveSimulationState($t$)
    \State $t \leftarrow t + \Delta t$
\EndWhile
\end{algorithmic}
\end{algorithm}

\section{Model Ensemble Configuration}

The generative processes within the urban mobility foundation model can be categorized into four types: planning, reactive behavior, chatting, and reflection. These tasks may be handled by different LLMs depending on their complexity and frequency, as summarized by Table~\ref{tab:llm generations}.

\refstepcounter{appendixbtable}
\label{tab:llm generations}

\begin{table*}
\centering
\textbf{Table \ref{tab:llm generations}.} Model Ensemble Configuration Summary.

{\footnotesize
\begin{tabular}{|p{3.6cm}|p{2.0cm}|p{5.5cm}|c|c|c|}
\hline
\textbf{[Type] Name} & \textbf{Vars in Prompt} & \textbf{Description | File} & \textbf{C} & \textbf{F} & \textbf{M} \\
\hline
[Plan] Initial Activity Plan Generation & 1, 2, 3, 4, 5, 6, 10, 11, 14 & Generate initial daily activity plan at the start of a day. \newline \texttt{generate\_daily\_activity\_plan} & 5 & 1 & L \\
\hline
[Reaction] Reaction Generation & 1, 2, 3, 4, 7, 8, 9, 10, 11, 12, 14 & Generate reactions to real-time network events and traffic state during a day. \newline \texttt{update\_daily\_activity\_plan} & 4 & 3 & M \\
\hline
[Reaction] Extracting Path Info & 17 & Utility tool to extract the path info. \newline \texttt{extract\_path\_info} & 1 & 3 & S \\
\hline
[Reflection] Reflection Generation & 1, 2, 3, 4, 7, 8 & Reflection generation at the end of a day. \newline \texttt{daily\_reflection} & 3 & 1 & M \\
\hline
\multirow{2}{*}{[Chat] Chat Initiation} 
& 1, 2, 3, 4, 10, 11 & Decisions regarding whether to initiate a conversation at the start of the day, and with whom to engage. \newline \texttt{initiate\_chat\_new\_day} & 1 & 1 & S \\
& 1, 2, 3, 4, 10, 11, 12 & Decisions regarding whether to initiate a conversation during the day, and with whom to engage. \newline \texttt{initiate\_chat\_during\_day} & 1 & 4 & S \\
\hline
[Chat] Response Generation & 1, 2, 3, 4, 10, 11, 12, 13 & Generate response to chat. \newline \texttt{generate\_response} & 1 & 4 & S \\
\hline
[Chat] Chat Summary & 1, 2, 3, 13 & Summarize chat interaction. \newline \texttt{generate\_chat\_summary} & 1 & 3 & S \\
\hline
[Memory] Importance Score Generation & 1, 2, 15, 16 & Generate importance score for network event, chat, or thought. \newline \texttt{generate\_importance\_score} & 1 & 5 & S \\
\hline
\end{tabular}
}

\vspace{1em}
\raggedright
{\footnotesize
Notes:

a) “C” indicates the complexity level of the generation task (1 to 5); 

b) “F” represents how frequently the generation is triggered (1 to 5); 

c) “M” suggests the recommended model size for the task, with possible values: large (L), medium (M), or small (S);

d) Meaning of indices in `Vars in Prompt`: 1) Simulation description; 2) Network description; 3) Person profile; 4) Curr time; 5) Prev day plan \& reflection; 6) Prev day reflection; 7) Today initial activity plan; 8) Today reaction history; 9) Current activity progress; 10) Perception 11) Retrieved; 12) Real-time traffic state; 13) Ongoing chat; 14) Recent chats; 15) Concept type; 16) Concept description; 17) Path string.
}

\end{table*}

\section{Psychologically-Informed Forgetting Mechanisms}

Realistic memory decay modeling prevents unbounded memory growth while reflecting psychological principles governing human memory retention. Each memory entry receives an expiration attribute determining its lifespan based on non-linear relationships with importance scores, reflecting the well-established principle that salient experiences persist longer than mundane ones \citep{anderson1991reflections, wixted1991form}. The importance score criteria is shared at \texttt{gatsim/technical\_highlights} \citep{liu2025gatsim}. The lifespan calculation employs power functions aligning with empirical findings on human memory retention:

\begin{equation}
\text{lifespan}_{\text{init}} = \text{lifespan}_{\text{min}} + (\text{lifespan}_{\text{max}} - \text{lifespan}_{\text{min}}) \times \text{importance}^{\text{power}}
\end{equation}

\begin{figure}
    \centering
    \includegraphics[width=0.7\linewidth]{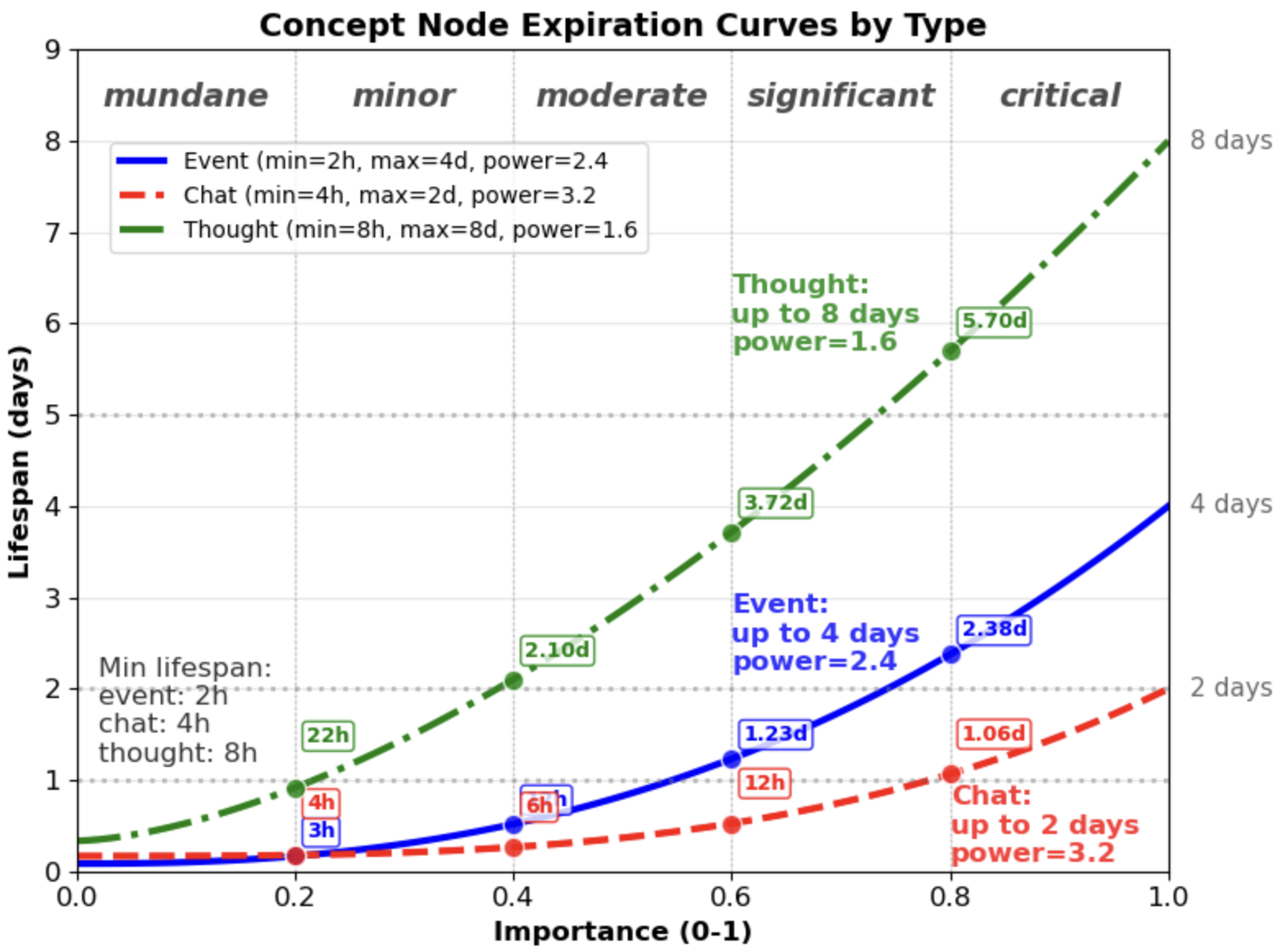}
    \caption{Concept expiration computation for different memory types. The figure illustrates how importance scores determine memory lifespans through power-law relationships, with different parameters for event memories, conversational memories, and reflective insights to reflect their distinct behavioral characteristics.}
    \label{fig:concept_expiration}
\end{figure}

In the prototype, parameters are tuned to each memory type's distinct characteristics. Event memories use minimum lifespans of 2 hours and maximum of 4 days with power 2.4, allowing transient irrelevant events to fade quickly while enabling important ones to persist for pattern learning. Conversational memories employ 4-hour minimums and 2-day maximums with power 3.2, reflecting that most coordination content are relevant but only pertains to immediate needs, while highly important exchanges still warrant extended retention. Reflective insights use 8-hour minimums and 8-day maximums with power 1.6, enabling learned knowledge to guide behavior over longer timescales (see Figure \ref{fig:concept_expiration}. This approach mirrors power-law forgetting curves observed in psychological studies \citep{wixted1991form} where retention probability follows $P(t) \propto t^{-\beta}$. Memory retrieval during reasoning or reflection extends expiration by the initial lifespan, implementing the spacing effect effect (the psychological phenomenon where repeated exposure to information over time strengthens memory retention) observed in human memory \citep{cepeda2006distributed}.

\section{Examples of Activity Continuity}

\refstepcounter{appendixdlisting}
\noindent
\label{list:example_cross_day_continuity}
\textbf{Listing \ref{list:example_cross_day_continuity}.} Example of activity continuity in GATSim: an agent maintains commitment to postponed activities through reflection and plan revision.
\vspace{0.5em}
\begin{lstlisting}[
  basicstyle=\ttfamily\footnotesize,   % smaller font
  lineskip=-1pt,                       % reduce line spacing
  aboveskip=4pt, belowskip=4pt,        % reduce vertical gaps before/after
  breaklines=true,
  escapeinside={(*}{*)},
]
Example of Activity Continuity

Day 1 (Tuesday):
- Initial Plan: Work -> Supermarket (17:30) -> Home
- During Day: Unexpected traffic incident causes agent to skip supermarket visit
- Daily Reflection: "I planned to visit the supermarket after work but had to skip it due to severe congestion on Ave_2. Need to reschedule this grocery shopping trip."

Day 2 (Wednesday) Morning:
- Context for planning: Last long-term reflection + yesterday's daily reflection
- Long-term Reflection Update: "Need to visit supermarket today - originally planned for Tuesday but postponed due to traffic disruption."
- Generated Plan: Work -> Supermarket (17:30) -> Home
- Reasoning: "Since I missed the supermarket visit yesterday due to traffic, I need to go today after work."

Day 2 (Wednesday) During Day:
- At each revision opportunity, context includes:
  * Long-term reflection (explaining why supermarket is important)
  * Today's initial plan (with supermarket visit)
  * Today's revision history
- This ensures the agent maintains commitment to important tasks throughout the day.
\end{lstlisting}

\section{Experiment Demographics}

The demographic composition reflects realistic urban patterns, with age distributions, household structures, employment categories, and vehicle ownership rates summarized in Table~\ref{tab:population_summary}.

\refstepcounter{appendixetable}
\label{tab:population_summary}

\begin{table}[h]
\textbf{Table \ref{tab:population_summary}.} Summary of Synthetic Population Characteristics.
\begin{minipage}[t]{0.35\textwidth}
\raggedright
\begin{tabular}{p{2cm}p{2cm}r}
\toprule
\textbf{Category} & \textbf{Subcategory} & \textbf{Count} \\
\midrule
\multirow{2}{2cm}{Gender} 
    & Female & 36 \\
    & Male & 34 \\
\midrule
\multirow{2}{2cm}{Age Group} 
    & Young & 60 \\
    & Children & 10 \\
\midrule
\multirow{5}{2cm}{Family Role} 
    & Single & 26 \\
    & Husband & 17 \\
    & Wife & 17 \\
    & Son & 5 \\
    & Daughter & 5 \\
\midrule
\multirow{5}{2cm}{Family Structure} 
    & Total families & 43 \\
    & 1-member & 26 \\
    & 2-member & 8 \\
    & 3-member & 8 \\
    & 4-member & 1 \\
\midrule
\multirow{3}{2cm}{Household Income} 
    & Low & 10 \\
    & Middle & 43 \\
    & High & 17 \\
\midrule
Licensed Drivers & & 50 \\
\bottomrule
\end{tabular}
\end{minipage}
\hspace{6em}
\begin{minipage}[t]{0.35\textwidth}
\raggedright
\begin{tabular}{p{2cm}p{2cm}r}
\toprule
\textbf{Category} & \textbf{Subcategory} & \textbf{Count} \\
\midrule
\multirow{3}{2cm}{Vehicles per Family} 
    & 0 vehicles & 4 \\
    & 1 vehicle & 25 \\
    & 2 vehicles & 14 \\
\midrule
\multirow{11}{2cm}{Workers per Facility} 
    & Coffee shop & 3 \\
    & Factory & 20 \\
    & Hospital & 3 \\
    & Gym & 3 \\
    & Office & 17 \\
    & Food court & 3 \\
    & Amusement park & 2 \\
    & Museum & 2 \\
    & Cinema & 2 \\
    & Supermarket & 3 \\
    & School & 2 \\
\midrule
\multirow{2}{2cm}{Residents} 
    & Uptown apartment & 36 \\
    & Midtown apartment & 34 \\
\bottomrule
\end{tabular}
\end{minipage}
\end{table}

\section{Examples of Generative Agent Reactive Behaviors and Reflections}

\refstepcounter{appendixflisting}
\noindent
\label{list:example_reactive_behaviors}
\textbf{Listing \ref{list:example_reactive_behaviors}.} Examples of generative agent flexible reactive behaviors.
\vspace{0.5em}
\begin{lstlisting}[
  basicstyle=\ttfamily\footnotesize,   % smaller font
  lineskip=-1pt,                       % reduce line spacing
  aboveskip=4pt, belowskip=4pt,        % reduce vertical gaps before/after
  breaklines=true,
  caption={},
]
Name Michael Chen | Time: Thu 2025-03-13 14:05
I am currently at Supermarket, having arrived at 12:40 for a planned 10-minute trip. However, I have been waiting for 80 minutes due to unforeseen queuing delays. This significantly impacts my schedule, particularly the upcoming meeting with Gloria Campbell at Office. Since there is no congestion in the road network and my travel time from Supermarket to Office is only 10 minutes by car, I can still make the originally scheduled 13:15 departure time if I leave now. Given my profession as a museum curator and the importance of finalizing exhibit details with Gloria, it's essential to maintain this meeting.
Plan revisions: Michael Chen update future activity plan:
[['Office', '14:10', 120, 'drive', 'Ave_2_link_3, St_4_link_1', 'Arrive at Office by 14:20 to meet with Gloria Campbell and finalize exhibit design details.'], ['Gym', '17:45', 60, 'transit', 'Metro_1', 'After work, take Metro_1 from Node_8 to Node_13 and transfer to visit Gym for light training and relaxation.'], ['Uptown apartment', '19:30', 'none', 'transit', 'Metro_2', 'Return home via Metro_2 to Node_1 and spend quiet evening with family before bedtime.']] 

Robert Johnson | Time: Thu 2025-03-13 07:01
I am currently en route to the Gym from Midtown apartment, but there is moderate congestion on St_1_link_2, where David Martinez is already experiencing a 5-minute delay. Although I have not yet encountered this link in my current trip, my planned path includes it, which could impact my punctuality. Given that my work as a fitness instructor requires me to be present early for morning classes, avoiding delays is crucial. Fortunately, alternative routes like using St_2_link_1 and Ave_2_link_2 are available, which bypass the congested area.
Plan revisions: Robert Johnson update current path to ['St_2_link_1', 'Ave_2_link_2'] 

Gloria Campbell | Time: Thu 2025-03-13 07:13
I am currently en route to work on the Factory, departing from Midtown apartment at 07:00. However, I have just encountered severe congestion on St_1_link_1 (Node_4 -> Node_9), with a wait time of 16 minutes reported. This will delay my arrival at the Factory unless I adjust my travel path. My original plan was to take St_1 and Ave_4; I can reroute efficiently to avoid further delays.
Plan revisions: Gloria Campbell update current path to ['Ave_2', 'St_4'] 

Christopher Evans | Time: Thu 2025-03-13 08:25
I am currently at the Coffee shop, where I have been discussing client training schedules with Benjamin Taylor. The current traffic state shows no congestion across any links in the network, which is a positive change from earlier delays on Ave_4_link_1 and St_1_link_2. This improved traffic condition allows for more flexibility in my upcoming travel plans. According to recent perception updates, Benjamin will remain at the Gym until at least 08:30 working with clients, so I plan to meet him there by 08:25 as previously coordinated. Since there is no ongoing congestion, I can proceed with confidence along the shortest available path to the Gym.
Plan revisions: Christopher Evans update current path to real-time shortest

Miguel Garcia | Time: Thu 2025-03-13 06:30
The traffic is currently clear with no congestion across the network, which means my usual route to work should be efficient. Based on yesterday's experience and previous reflections, I've learned that Ave_3_link_3 and St_1_link_2 can cause delays during evening commutes. To avoid this, I've coordinated with Nina Scott to reschedule our workout session for today at 17:30 at the Gym. I will leave work by 17:15 to ensure timely arrival and have planned an alternate route using Ave_1 and St_3 to bypass potential congestion.
Plan revisions: Miguel Garcia update future activity plan:
[['Amusement park', '08:00', 120, 'drive', 'St_2_link_2, Ave_1_link_1, St_2_link_1', "Drive to the Amusement Park to start a day's work as manager."], ['Food court', '12:30', 20, 'drive', 'shortest', 'Head to Food Court for lunch break after completing midday inspections.'], ['Gym', '17:30', 60, 'drive', 'Ave_1, St_3_link_2', 'Go to Gym for an evening workout session with Nina Scott before heading back home.'], ['Uptown apartment', '19:30', 'none', 'drive', 'shortest', 'Wrap up his day by driving back home to his Uptown apartment, preparing for a restful night.']] 
\end{lstlisting}

\refstepcounter{appendixflisting}
\noindent
\label{list:example_reflections}
\textbf{Listing \ref{list:example_reflections}.} Examples of generative agent gaining knowledge through reflections on travel experiences.
\vspace{0.5em}
\begin{lstlisting}[
  basicstyle=\ttfamily\footnotesize,   % smaller font
  lineskip=-1pt,                       % reduce line spacing
  aboveskip=4pt, belowskip=4pt,        % reduce vertical gaps before/after
  breaklines=true,
  caption={},
]
Name: Rachel Thompson | Date: Thu 2025-03-13 AM
This week has been a balance of managing the Cinema's operations and coordinating community events with Isabella Rodriguez. Yesterday's traffic disruptions on St_1_link_2 highlighted the need to proactively adjust evening routes, especially when coordinating with Marcus for post-work plans.

Hannah Wilson | Date: Thu 2025-03-13 AM
Traffic was mostly clear throughout the day, with only minor delays on Ave_4_link_1 and St_1_link_2. Today, I need to maintain flexibility while ensuring punctuality for work and social commitments. I'll prioritize leaving early enough to avoid potential congestion on St_3_link_2 and ensure smooth coordination with Brian and Frank for post-work activities.

Patricia Johnson | Date: Thu 2025-03-13 AM
The past week has been a balance of work, family coordination, and personal commitments. I've successfully adjusted my commute routes to avoid congestion on St_1_link_2 and St_1_link_1, which has improved my punctuality at the Factory.

Michelle Harris  | Date: Thu 2025-03-13 AM
I've learned to adjust my routines to avoid traffic bottlenecks by leaving earlier and using alternative routes like Ave_2 during morning commutes. The evening meetup with Elizabeth at Coffee shop worked well as an alternative to the Gym, especially since the Gym tends to be crowded before 20:30. 

Raymond Foster  | Date: Thu 2025-03-13 PM
Overall, today highlighted the importance of adaptive routing during peak hours and planning buffer time for social activities that may overrun. I learned that St_1_link_2 and Ave_2_link_1 consistently experience congestion in the morning, and Ave_4_link_1 becomes problematic in the evening. Future planning should include earlier departures and alternative routes during these windows.
\end{lstlisting}

\section{Examples of Identified Limitations}

\refstepcounter{appendixglisting}
\noindent
\label{list:problem_hallucination}
\textbf{Listing \ref{list:problem_hallucination}.} Examples of hallucination problem.
\vspace{0.5em}
\begin{lstlisting}[
  basicstyle=\ttfamily\small,
  breaklines=true,
  escapeinside={(*}{*)},
  caption={}
]
Hallucinated travel experiences on the first day of simulation:
The past week was busy with work at the coffee shop, but I managed to maintain a good routine of morning yoga
and early starts. Traffic on Ave_3 has been consistently congested during the morning commute, so I've adjusted
my departure times slightly earlier to avoid delays. This week, I'm looking forward to spending more time with
Sophia and exploring the Museum's modern art exhibition. I also need to ensure I leave work on time to meet
Sophia at the Museum by 12:30. After the exhibition, lunch at the cafe sounds like a nice way to relax.

Hallucinated link (Ave_3_link_3) travel experience:
Reflection: This week has been quite productive so far. I've managed to maintain a smooth commute to the Amusement Park and have coordinated well with Ashley regarding car usage, which has made the mornings stress-free. My route via St_2_link_1 and Ave_3_link_3 continues to be reliable in the morning if I leave before 07:00, and yesterday's traffic was lighter than expected on the return trip. 
\end{lstlisting}

\refstepcounter{appendixglisting}
\noindent
\label{list:problem_instruction_following}
\textbf{Listing \ref{list:problem_instruction_following}.} Examples of instruction following problem.
\vspace{0.5em}
\begin{lstlisting}[
  basicstyle=\ttfamily\footnotesize,   % smaller font
  lineskip=-1pt,                       % reduce line spacing
  aboveskip=4pt, belowskip=4pt,        % reduce vertical gaps before/after
  breaklines=true,
  escapeinside={(*}{*)},
  caption={}
]
Repeated chatting the same topic with the same person:
Summary of Isabella Rodriguez's chats with other people:
---CHAT SUMMARY SECTION START---
Isabella Rodriguez and Sophia Nguyen decided to watch a movie at the Cinema instead of their usual plans. They agreed to attend the 3:00 PM showing and will leave the Coffee shop by 2:30 PM to meet at the Cinema.
---CHAT SUMMARY SECTION END---
Now generate your JSON chat request:
output
'{
    "person_name": "Sophia Nguyen",\n (*\ul{The new Captain America movie is showing at the Cinema today. Want to catch the 3:00 PM show as planned?}*)
    "query": " I\'ll leave the Coffee shop by 2:30 PM."
}'

Generating inconsistent vehicle usage:
[['Midtown apartment', '06:15', 20, 'none', 'none', 'Wake up at home and complete morning routines including light stretching and breakfast preparation.'], ['Factory', '06:45', 120, 'transit', 'Metro_2_link_1, Metro_2_link_2, Metro_2_link_3', 'Take Metro_2 with Christopher Evans to Factory for an 8:00 start time, departing earlier than usual to test alternative transit route.'], (*\ul{['Supermarket', '17:10', 20, 'drive', 'shortest', 'Head to Supermarket during lunch break to stock up on groceries for the upcoming days.']}*), ['Gym', '17:30', 60, 'transit', 'Metro_1_link_1, Metro_1_link_2, Metro_1_link_3, Metro_1_link_4', 'Take Metro_1 to Gym for a strength training session with Christopher Evans starting at 17:00.'], ['Midtown apartment', '19:00', 'none', 'transit', 'Metro_1_link_4, Metro_1_link_3, Metro_1_link_2, Metro_1_link_1', 'Return home via Metro_1 after completing the workout session.']]
\end{lstlisting}

\refstepcounter{appendixglisting}
\noindent
\label{list:problem_commensense}
\textbf{Listing \ref{list:problem_commensense}.} Examples of commonsense violation.
\vspace{0.5em}
\begin{lstlisting}[
  basicstyle=\ttfamily\footnotesize,   % smaller font
  lineskip=-1pt,                       % reduce line spacing
  aboveskip=4pt, belowskip=4pt,        % reduce vertical gaps before/after
  breaklines=true,
  escapeinside={(*}{*)},
  caption={}
]
Plan to eat lunch at 19:00:
['Food court', '19:00', 'none', 'drive', 'shortest', 'Drive to the Food court to meet Isabella Rodriguez (*\ul{for lunch after the workout}*).']

The agent plans activities for times that have already passed:
(Chat history between Isabella Rodriguez and Sophia Nguyen)
Isabella Rodriguez: The new Marvel movie is showing at the Cinema starting 10:00 AM today. Should we catch it after the art exhibition at the Museum? It might be busy, so we should leave early to avoid traffic. 
Sophia Nguyen: That sounds fun! Let's aim to leave the Museum by 9:30 AM to give us enough time to get to the Cinema without rushing. 
Isabella Rodriguez: Great! (*\ul{Let's meet at the Museum entrance by 11:30 AM, then leave together for the Cinema around 9:30 AM.}*). I'll drive us there to save time. 
Sophia Nguyen: Sounds good! I'll meet you at the Museum entrance by 11:30 AM. Let's confirm closer to the time if anything changes. Isabella Rodriguez: Will do! I'll see you at the Museum entrance at 11:30 AM. Safe travels!
\end{lstlisting}

\refstepcounter{appendixglisting}
\noindent
\label{list:problem_reasoning}
\textbf{Listing \ref{list:problem_reasoning}.} Examples of reasoning deficits.
\vspace{0.5em}
\begin{lstlisting}[
  basicstyle=\ttfamily\footnotesize,   % smaller font
  lineskip=-1pt,                       % reduce line spacing
  aboveskip=4pt, belowskip=4pt,        % reduce vertical gaps before/after
  breaklines=true,
  escapeinside={(*}{*)},
  caption={}
]
Displays concern about congestion unrelated to the current travel context:
I am currently en route to the Museum after departing my Uptown apartment at 07:15. (*\ul{Let's meet at the Museum entrance by 11:30 AM, then leave together for the Cinema around 9:30 AM.}*) The current traffic state shows moderate congestion on St_2_link_1 and light congestion on Ave_2_link_1, but these do not affect my current travel path. Upon arrival at the Museum around 07:33, I will proceed with collaborative research with Gloria Campbell before returning to the Coffee shop for my managerial duties starting at 10:00 AM.

Insufficient time allocated for the round trip to Museum during the lunch break:
[['Midtown apartment', '06:00', '20', 'none', 'none', 'Wake up at home, complete morning routine.'],
['Office', '07:15', 'none', 'drive', 'Ave_2, St_5', 'Drive to the Office to start work.'], 
(*\ul{['Museum', '12:00', 'none', 'drive', 'shortest', 'Meet Sophia Nguyen at the Museum for the modern art exhibition.'],}*)
['Office', '13:00', '120', 'drive', 'shortest', 'Return to the Office to continue working.'],
['Midtown apartment', '19:30', 'none', 'drive', 'shortest', 'Drive back home to rest and prepare for the next day.']]

Generate invalid path:
I encountered severe congestion on St_1_link_1 earlier this morning, which required me to reroute via Ave_2 and St_3 to reach the Factory. Currently, I am traveling on St_1_link_2 toward Node_13 (Factory) and experiencing a moderate delay with 5 minutes of wait time. Despite these delays, I expect to arrive at the Factory before 8:00 as planned. My previous communication with Catherine Young and Oliver Green confirms that taking Ave_4_link_1 and St_4_link_2 is an effective alternative path to avoid further congestion.
(*\ul{Plan revisions: Gloria Campbell update current path to 'Ave\_4', 'St\_4'}*)

Wrong reasoning - new path cannot avoid link St_4_link_2:
At 08:32, I am en route to the Factory and currently at Node_7 after taking Ave_1_link_1, St_3_link_1, and Ave_2_link_3. My current planned path includes St_4_link_1 -> St_4_link_2 to reach the Factory. However, based on past experience and the concepts retrieved, I recall that St_4_link_2 has a history of moderate congestion during the morning commute, which could impact my punctuality. (*\ul{Even though real-time traffic shows no congestion, proactively avoiding historically problematic segments like St\_4\_link\_2 aligns with my previous adjustments and helps ensure I arrive at work on time without unnecessary stress.}*)
\end{lstlisting}

\refstepcounter{appendixglisting}
\noindent
\label{list:problem_multiagent_coordination}
\textbf{Listing \ref{list:problem_multiagent_coordination}.} Examples of multi-agent coordination failure.
\vspace{0.5em}
\begin{lstlisting}[
  basicstyle=\ttfamily\footnotesize,   % smaller font
  lineskip=-1pt,                       % reduce line spacing
  aboveskip=4pt, belowskip=4pt,        % reduce vertical gaps before/after
  breaklines=true,
  escapeinside={(*}{*)},
  caption={}
]
Dupicated picking-up activities:

(Husband plan)
Uptown apartment, 06:45 (none) - Wake up at home and complete morning routines while having breakfast with Sarah and Emily.
School, 07:20 (drive) - Drive Emily to School and drop her off before heading to the Coffee shop.
Coffee shop, none (drive) - Grab coffee and light snack at the Coffee shop before heading to work.
Museum, 08:15 (drive) - Arrive at Museum by 8:30 to start work and prepare for the modern art exhibition setup.
Supermarket, 12:00 (drive) - Take a short break to buy groceries for the family before returning to work.
Museum,  13:15 (drive) - Resume work at the Museum and oversee the launch of the modern art exhibition at 12:00 PM.
(*\ul{School, 16:45 (drive) - Drive back to School to pick up Emily after finishing key tasks at work.}*)
Food court, none (drive) - Have dinner together at Food court before heading back home.
Uptown apartment, 19:00 (drive) - Return home to Uptown apartment by 19:30 for a quiet evening with family.

(Wife Plan)
Uptown apartment, 06:00 (none) - Wake up at Uptown apartment and complete morning hygiene routines.
Gym, 06:45 (drive) - Drive to Gym for a morning yoga session with Isabella Rodriguez.
Coffee shop, none (drive) - Head to Coffee shop from Gym with Isabella for coffee and breakfast before work.
Hospital, none (drive) - Arrive at Hospital by 8:00 AM to begin scheduled clinical duties (8:00 - 11:30, 13:30 - 17:00).
(*\ul{School, 15:15 (drive) - Leave Hospital at 15:10 to pick up daughter Emily Chen from School.}*)
Uptown apartment, 17:30 (drive) - Return home to Uptown apartment, prepare dinner, and unwind after a long day.
\end{lstlisting}

\end{document}